\documentclass{article} 
\usepackage{iclr2027_arxiv,times}

\usepackage{amsmath,amsfonts,bm}

\def\eqref#1{equation~\ref{#1}}

\def\1{\bm{1}}

\DeclareMathAlphabet{\mathsfit}{\encodingdefault}{\sfdefault}{m}{sl}
\SetMathAlphabet{\mathsfit}{bold}{\encodingdefault}{\sfdefault}{bx}{n}

\usepackage{hyperref}
\usepackage{url}

\usepackage{graphicx}
\usepackage{wrapfig}
\usepackage{booktabs}
\usepackage{multirow}
\usepackage[table]{xcolor}
\usepackage{pifont}
\newcommand{\xmark}{\ding{55}}
\usepackage{adjustbox}
\usepackage{tcolorbox}
\newcounter{prompt}

\tcbset{promptstyle/.style={colback=gray!5, colframe=gray!75!black, fonttitle=\bfseries\small,
  left=4pt, right=4pt, top=3pt, bottom=3pt}}
  
\title{Latent World Action Model for GUI Agents}

\author{Seungjun Yu$^{*}$,\ Hojun Choi$^{*}$,\ Seojeong Park,\ Jaeyo Shin,\ Hyunjung Shim$^{\dagger}$\\
$^{*}$Equal contribution. \quad $^{\dagger}$Corresponding author.\\
Korea Advanced Institute of Science and Technology (KAIST)\\
\texttt{\{seungjunyu,hchoi256,seojeong.park,jaeyo\_shin,kateshim\}@kaist.ac.kr}
}

\usepackage{xcolor}
\usepackage{titletoc}

\definecolor{appendixblue}{rgb}{0.21,0.49,0.74}
  
\iclrfinalcopy 
\begin{document}

\maketitle

\begingroup
\renewcommand{\thefootnote}{}
\footnotetext{Project page: \url{https://hchoi256.github.io/projects/latent2world/}}
\endgroup

\begin{abstract}
Pixel-space world models have recently emerged as a promising approach for GUI agents, predicting future screens prior to execution and subsequently re-perceiving them for action selection. However, this two-stage paradigm faces two key challenges in GUI environments: (L1) the high computational cost of pixel reconstruction and (L2) the need to reconstruct large and diverse appearance-level details beyond what is required for action selection. In practice, this screen-generation objective can induce an identity shortcut problem, where future-state prediction largely preserves the current state rather than capturing action-induced changes. In this work, we reformulate GUI world modeling by shifting the prediction target from full-screen reconstruction to action-induced future information in latent space. Concretely, we leverage a large multimodal model as the latent world model, whose shared representations support two dedicated heads for both action selection and predictive transition modeling. The action head directly translates these latent representations into action selection, while the transition head captures action-conditioned future information. To mitigate the identity shortcut, we further introduce an action-aware contrastive objective that encourages the transition representations to capture discriminative action-induced changes. Extensive experiments across offline and online GUI settings show that our method achieves up to 31.9\% and 40.6\% relative gains in success rate over pixel-space world models, respectively, while substantially reducing inference latency by up to 98.9\%.
Our code is available at \url{https://github.com/sjyu001/latent2world}.
\end{abstract}

\section{Introduction}
\label{sec:intro}

Recent advancements in large multimodal models (LMMs) have accelerated the development of graphical user interface (GUI) agents~\citep{cogagent,androidinthewild,mobilegpt,tora}---an increasingly indispensable technology in the modern digital era. These agents are designed to complete complex, user-specified tasks on mobile platforms by automatically executing sequences of actions. By reducing manual interaction, they facilitate mobile app use across diverse real-world domains, ranging from commercial services such as e-commerce~\citep{appagent} to accessibility support for people with visual or motor impairments and older adults~\citep{syed}.

Among such efforts, recent works~\citep{xu} have explored world models that simulate the consequences of actions proposed by GUI agents (\emph{e.g.}, button clicks and scrolling) prior to direct execution. Such foresight enables GUI agents to anticipate potential failures and adjust their strategies accordingly, improving robustness in long-tail and irreversible scenarios (\emph{e.g.}, payments and account deletion), as well as in long-horizon tasks involving extended action sequences. One promising approach is visual-space prediction~\citep{code2world,vimo}, where GUI world models forecast a corresponding next GUI state for each candidate action in a visually renderable form (\emph{e.g.}, a screenshot or renderable HTML). These predicted future states are then re-perceived by a separate action-scoring model, which selects the candidate that best advances the task goal. This two-stage ``Render-and-Re-Perceive'' pipeline provides explicit spatial and visual information, such as element locations and layouts, that can support grounded GUI action planning beyond relatively concise textual descriptions~\citep{gu}.


\begin{wrapfigure}{r}{0.55\textwidth}
    \centering
    \includegraphics[width=\linewidth]{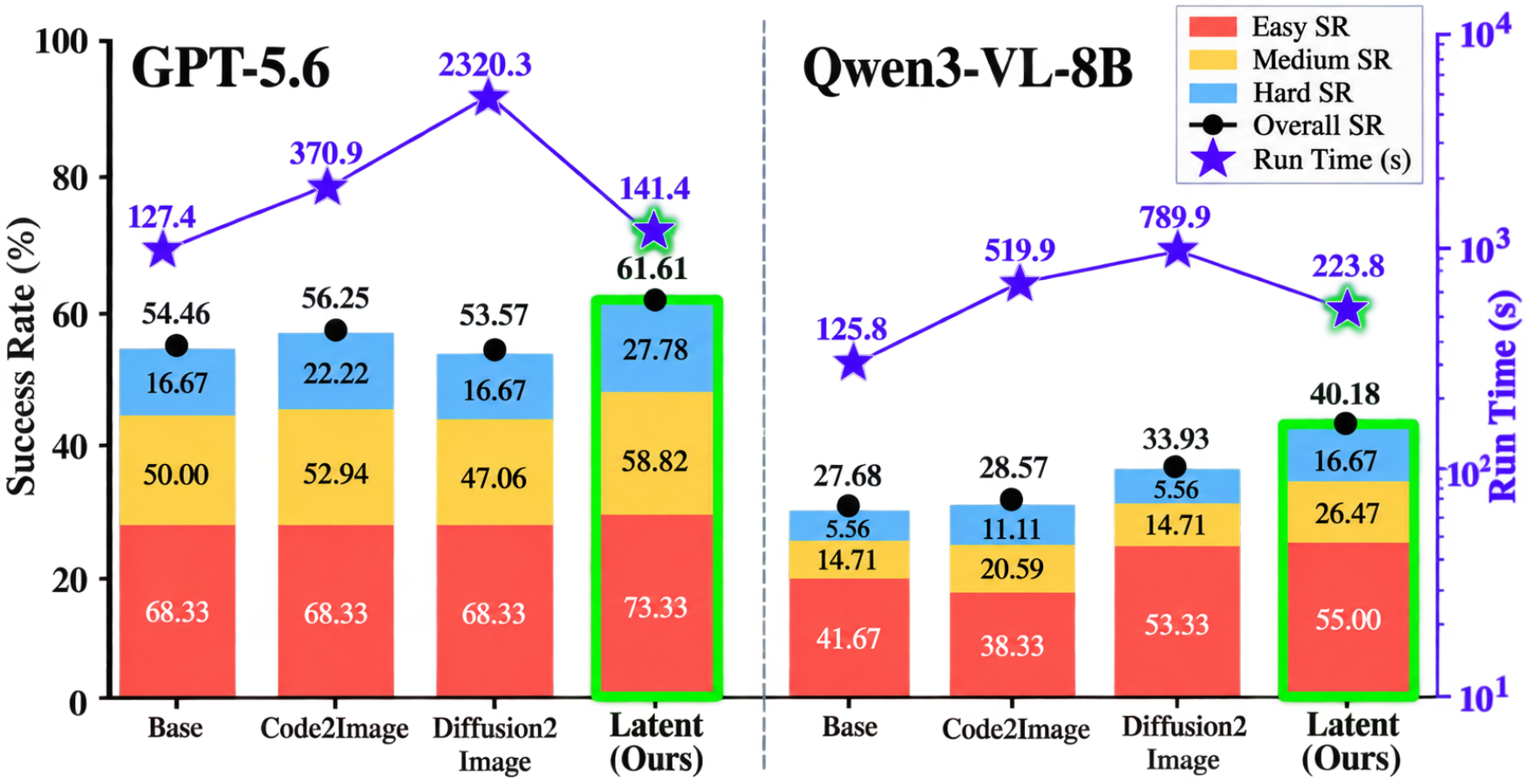}
    \vspace{-6mm}
    \caption{Our method, leveraging latent future-state representations, achieves the best GUI navigation performance with substantially improved efficiency.}
    \label{fig:f1}
\end{wrapfigure}

\begin{wrapfigure}{r}{0.57\textwidth}
    \centering
    \includegraphics[width=\linewidth]{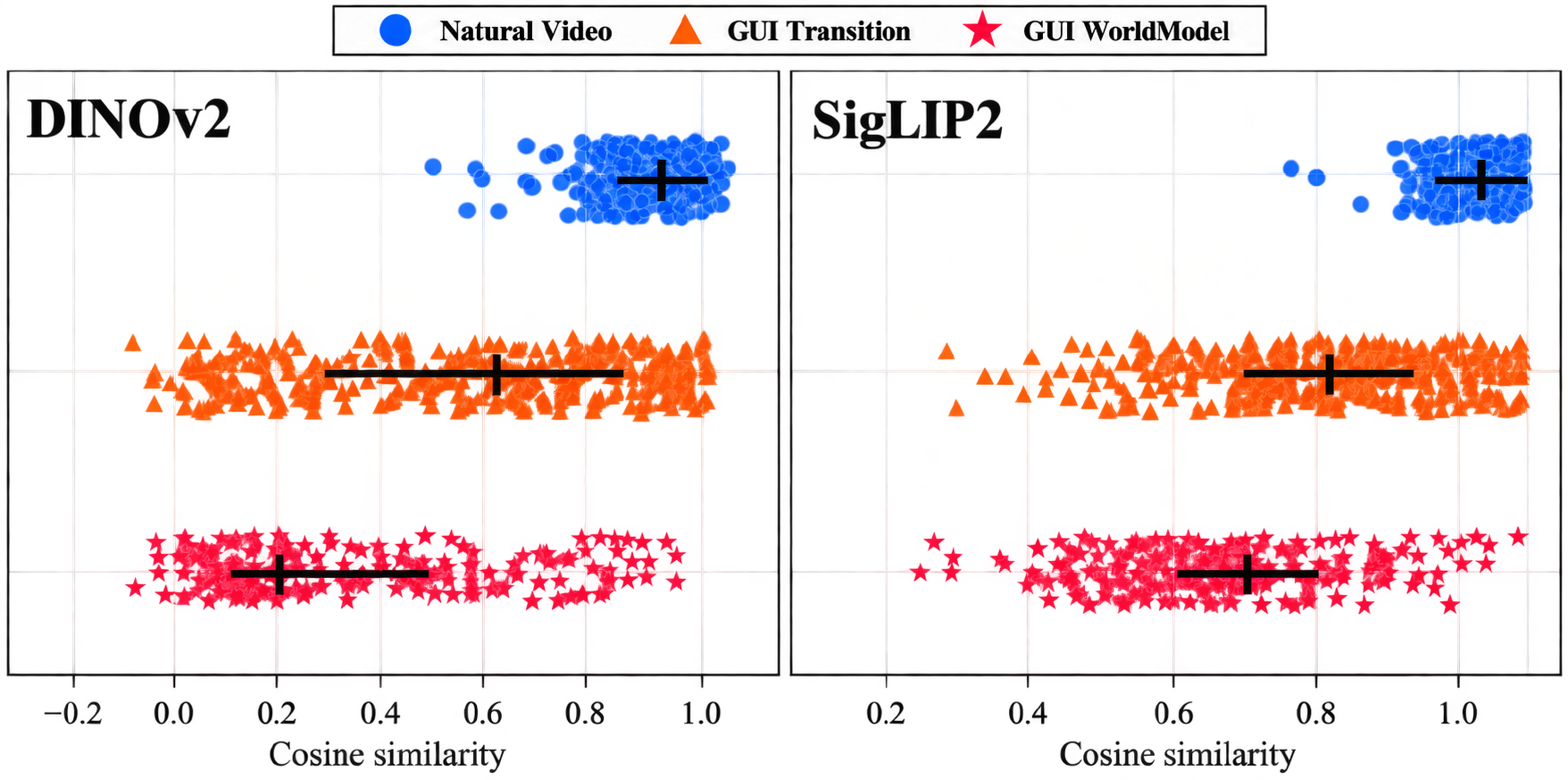}
    \vspace{-6mm}
    \caption{Natural videos exhibit strong temporal continuity, whereas GUI transitions show lower visual similarity and large variance, increasing the range of future appearance variations that must be modeled in pixel space.}
    \label{fig:f2}
\end{wrapfigure}

However, explicitly reconstructing future screens introduces an expensive and unnecessarily broad intermediate task into action selection. \textbf{(L1)} Autoregressive or diffusion-based screen synthesis incurs substantial computational latency, posing a bottleneck for responsive GUI interaction (Figure~\ref{fig:f1}). Moreover, \textbf{(L2)} GUI transitions exhibit large and varied appearance changes (Figure~\ref{fig:f2}), requiring pixel-space world models to capture diverse future appearances. Yet, action selection does not require reconstructing all such visual details, but rather understanding the future consequence of each action. 
In practice, pixel-space world models may retain the current screen despite substantial next-screen changes (Figure~\ref{fig:f4}), a behavior we term an \textit{identity shortcut}. 
We hypothesize that full-screen reconstruction encourages reusing current-screen information instead of modeling action-induced changes.
This distinction raises the question: \textit{how can a GUI world model learn future information useful for action selection without reconstructing the future screen?}

Our key insight is that GUI world modeling should focus on future information that helps distinguish which candidate action best advances the task goal. This perspective requires connecting what an action will cause with whether that outcome is useful for the task. Training on annotated actions teaches a model which action to select, but does not explicitly require it to anticipate the resulting state. Conversely, learning to predict observed next states provides information about action consequences, but does not directly teach the model how to use that information for action selection. We therefore jointly train shared representations for both candidate-action scoring and action-conditioned future-state prediction. The predictive objective encourages these representations to capture action-induced future information, while the action-selection objective connects them to task-relevant decisions. Accordingly, we implement the predictive objective in latent space, enabling the model to learn from future states without reconstructing their full visual appearance.

Specifically, given the task goal, current GUI screenshot, action history, and candidate actions proposed by a frozen GUI agent, we employ a LoRA-adapted LMM to produce action-specific hidden representations. A lightweight action head maps each representation to a scalar score, and the resulting candidate distribution is trained against the ground-truth one-hot distribution identifying the annotated action. In parallel, a transition head predicts a next-state visual embedding from the shared representation of the observed action, using an embedding of the actual next screenshot as its target. However, simply minimizing the distance to this target may allow the previously observed \textit{identity shortcut}. We therefore train the transition head with an action-aware contrastive objective, aligning its prediction with the next-state embedding while separating it from the current-state embedding and embeddings of unrelated transitions. Both objectives jointly update the shared backbone, allowing future-state supervision to shape the representations used for action scoring. At inference time, the transition head is discarded, and the shared backbone and action head evaluate candidate actions in a single forward pass. This design bypasses the conventional ``Render-and-Re-Perceive'' pipeline while using representations trained to support both action selection and future-state prediction.

Through extensive experiments in complementary offline and online settings~\citep{androidcontrol,androidworld}, our method improves success rates over the corresponding base agents that do not use world modeling by up to 11.6\% and 47.0\%, respectively. It further outperforms pixel-space world models by up to 31.9\% and 40.6\%, while reducing world-model inference latency by up to 98.9\%.
\begin{itemize}
    \item We revisit GUI world modeling by shifting the prediction target from full-screen reconstruction to action-induced future information for effective action selection in latent space.
    \item We present a latent GUI world-action model that jointly learns action selection and action-conditioned future prediction with action-aware contrastive learning over latent transitions.
    \item Experiments show performance gains over both pixel-space world models and base GUI agents, while substantially reducing inference latency relative to pixel-space world models.
\end{itemize}

\section{Related Works}
\label{sec:related}

\subsection{Mobile GUI Agents}
Recent advances in LMMs have substantially improved mobile GUI agents, shifting interaction from metadata- or text-centric interfaces toward direct visual perception and grounded control~\citep{cogagent,showui,uitars}. Given a user instruction and the current GUI observation, these agents perform human-like operations such as tapping, typing, and swiping to accomplish complex tasks across mobile applications~\citep{mobileagent}. Subsequent studies have further enhanced their capabilities through improved visual grounding~\citep{uivision}, interaction-history modeling and structured reasoning~\citep{guirise}, self-reflection and error correction~\citep{guireflection}, and reinforcement-learning-based policy optimization~\citep{mobileguirl,arpo}. Building on this progress, mobile GUI agents have demonstrated broad practical value by automating routine interactions in commercial applications such as e-commerce~\citep{appagent} and reducing interaction barriers for users with visual or motor impairments and older adults~\citep{screen,syed}. However, these advances primarily improve the action-selection capabilities of GUI agents through better perception, reasoning, and policy optimization. Our work instead focuses on the complementary question of how predictive world-model information should be represented and integrated within GUI agents to explicitly better support action selection.

\subsection{GUI World Model for App Agents}
GUI world models provide agents with predictive knowledge of environment dynamics by simulating action-conditioned future states before direct execution. Early approaches primarily operate in textual space, predicting concise natural-language descriptions of state changes~\citep{chae} or imagined outcomes of candidate actions for model-based planning~\citep{gu}. Such textual abstractions efficiently capture semantic transitions but provide limited spatial and visual details. To retain more task-relevant GUI structure, subsequent work represents future states as textual sketches containing semantic and spatial information~\citep{mobiledreamer}. A complementary line moves toward explicit visual reconstruction. Diffusion-based approaches~\citep{vimo,xu} synthesize future GUI screenshots with textual information, while renderable-code-based methods~\citep{code2world} predict HTML combining structured interface information with deterministic visual rendering. However, explicitly materializing future GUI states requires modeling broad appearance-level details, incurring computational cost beyond what action selection requires. Latent predictive modeling provides an alternative and has been explored for action modeling in embodied settings~\citep{lawam}. Motivated by this mismatch, our work brings latent future prediction to GUI action selection. We shift the prediction target from full-state reconstruction to action-induced future information, enabling more efficient use of predictive knowledge for action selection.

\section{Methodology}
\label{sec:method}
In this work, we reformulate GUI world modeling by shifting the prediction target from full-screen reconstruction to action-induced future information in latent space. Our approach learns shared representations that jointly support complementary objectives: action selection identifies which candidate action best advances the task goal, while predictive world modeling captures future consequence induced by that action. This formulation directly integrates future-state predictive capability into action selection while avoiding unnecessary visual reconstruction and computational overhead.

We first introduce the latent world-action model, which employs a LoRA-adapted large multimodal model (LMM) to produce action-specific latent representations shared across world modeling and action selection. These representations are directly mapped to candidate-action scores for task-level decision making (Section~\ref{sec:method:m1}). We then formulate predictive latent modeling of GUI transitions, where a transition head predicts action-conditioned future representations from the shared latent features without explicit screen reconstruction. To capture action-induced changes in these predictive representations, we further present an action-aware contrastive objective (Section~\ref{sec:method:m2}). The resulting world-action model is jointly optimized during training while enabling efficient inference.

\subsection{Latent World-Action Model for GUI Agents}
\label{sec:method:m1}
Our formulation is motivated by the observation that action selection does not require reconstructing the full visual appearance of a future GUI state. Instead, a world model should preserve the future information necessary to evaluate the consequence of each action. This consideration is particularly relevant in GUI environments, where even a single atomic action can induce substantial appearance changes, while action evaluation may depend on only a small subset of the resulting information. Such a setting naturally motivates latent representations, which provide compact semantic abstractions and have been successfully adopted for predictive modeling across robotics~\citep{univla}, autonomous driving~\citep{latentcot,driveworldvla}, and video generation~\citep{drivelaw}. For example, if selecting the ``Send'' button leads to a login page, recognizing the resulting authentication requirement can be sufficient for evaluating the action without reproducing the entire screen.

\begin{figure}[!t]
    \centering
    \includegraphics[width=0.98\linewidth]{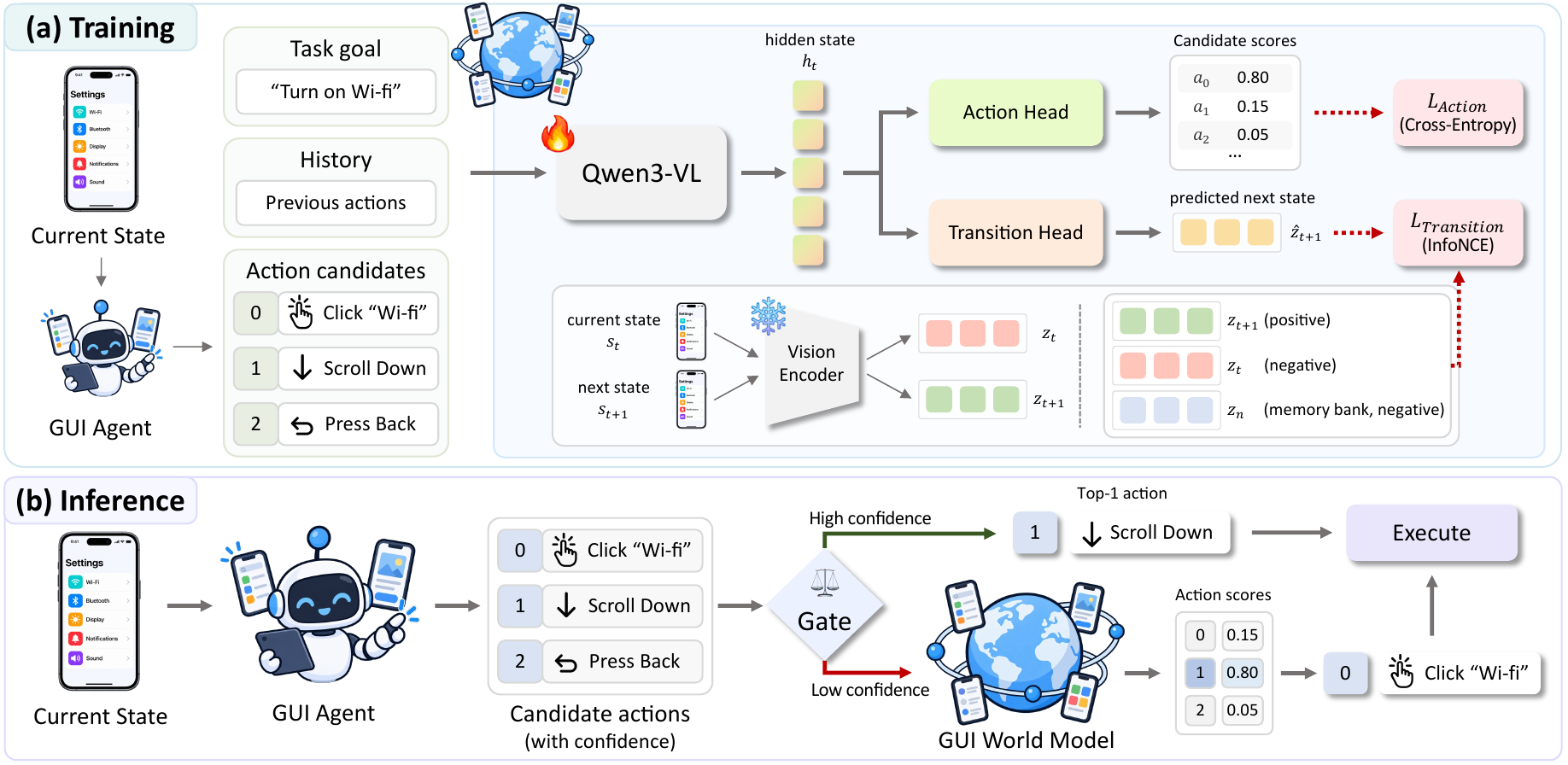}
    \vspace{-3mm}
    \caption{\textbf{Overview of our latent world-action model for GUI agents.} During training, we leverage a large multimodal model to learn action-specific shared representations through action scoring and contrastive future prediction. At inference, the transition head is removed for single-pass action scoring, while a confidence gate on the agent determines whether to invoke world-model rescoring.}
    \label{fig:f3}
    \vspace{-4mm}
\end{figure}

Building on this insight, we introduce a latent world-action model that uses the hidden representations of an LMM as a shared representation space for world modeling and action selection (Figure~\ref{fig:f3}). Given the current GUI context and candidate actions, the model produces an action-specific latent representation for each candidate. A lightweight action head then maps these representations to candidate-action scores, while a transition head operates on the same latent space to predict action-conditioned future representations. This shared formulation allows predictive supervision to directly shape the representations used for action selection without explicit screen reconstruction.

\subsubsection{Latent Representations for GUI Action Selection}
\label{sec:method:m11}
We first instantiate the action-selection branch of our latent world-action model by directly scoring candidate actions from LMM hidden representations, avoiding explicit visual decoding. At each interaction step, given the current GUI screenshot $s_t$, task goal $g$, interaction history $\mathcal{H}_t$, and a set of $K$ candidate actions $\mathcal{A}_t=\{a_t^1,\dots,a_t^K\}$ proposed by a frozen GUI agent, our LoRA-adapted model $\mathcal{W}_{\theta}$~\citep{qwen3vl} encodes all candidates jointly into a shared latent representation:
\begin{equation}
    h_t
    =
    \mathcal{W}_{\theta}
    \big(s_t,g,\mathcal{H}_t,\mathcal{A}_t\big),
\end{equation}
where the candidates are placed in a single prompt in a randomly permuted order and $h_t$ is the last-layer hidden state at the final prompt token. These representations encode candidate-specific information for downstream action selection directly within the shared predictive latent space. A lightweight action head then maps each $h_t^k$ to a scalar score, yielding the candidate-action distribution $p_t$. We supervise this distribution with the one-hot ground-truth target $y_t$ using cross-entropy:
\begin{equation}
    p_t
    =
    \mathrm{softmax}\!\left(
    \left[
    \operatorname{ActionHead}(h_t)_k
    \right]_{k=1}^{K}
    \right),
    \qquad
    \mathcal{L}_{\mathrm{Action}}
    =
    -\sum_{k=1}^{K} y_t^k \log p_t^k,
    \label{eq:act_loss}
\end{equation}
where $\operatorname{ActionHead}(h_t)_k = \mathbf{w}_k^{\top} h_t$ and $\mathbf{w}_k$ is the output embedding of the $k$-th index token.
This formulation directly maps shared representations to action decisions without reconstructing future screens. We next extend this latent space to predictive GUI transition modeling in Section~\ref{sec:method:m2}.

\subsection{Predictive Latent Modeling of GUI Transitions}
\label{sec:method:m2}
While the action-selection objective introduced above trains the shared representations to distinguish among candidate actions, it does not explicitly require them to capture the future consequence induced by each action. We therefore augment action supervision with predictive modeling of GUI transitions as an auxiliary training objective within the same LMM latent space. Rather than reconstructing the full visual appearance of the next screen, we train the model to capture action-conditioned future information that can directly shape the representations used for action selection.

Concretely, a lightweight transition head maps the same shared representation $h_t$ to the latent representation of the next state reached by the annotated action within this shared space to provide predictive supervision, while the actual next screenshot $s_{t+1}$ is encoded by a target encoder $\bar{\phi}$:
\begin{equation}
    \hat{z}_{t+1}
    =
    \operatorname{TransitionHead}(h_t),
    \qquad
    z_{t+1}
    =
    \bar{\phi}(s_{t+1}).
    \label{eq:transition_pred}
\end{equation}
Because $h_t$ is also the representation used to score the candidates, and their order is permuted in every example, predicting the next state of the annotated action requires $h_t$ to encode which candidate is correct and what it leads to. This predictive branch therefore provides complementary future-state supervision directly in the shared latent space without explicitly reconstructing the next screen.

\begin{figure}[!t]
    \centering
    \includegraphics[width=0.98\linewidth]{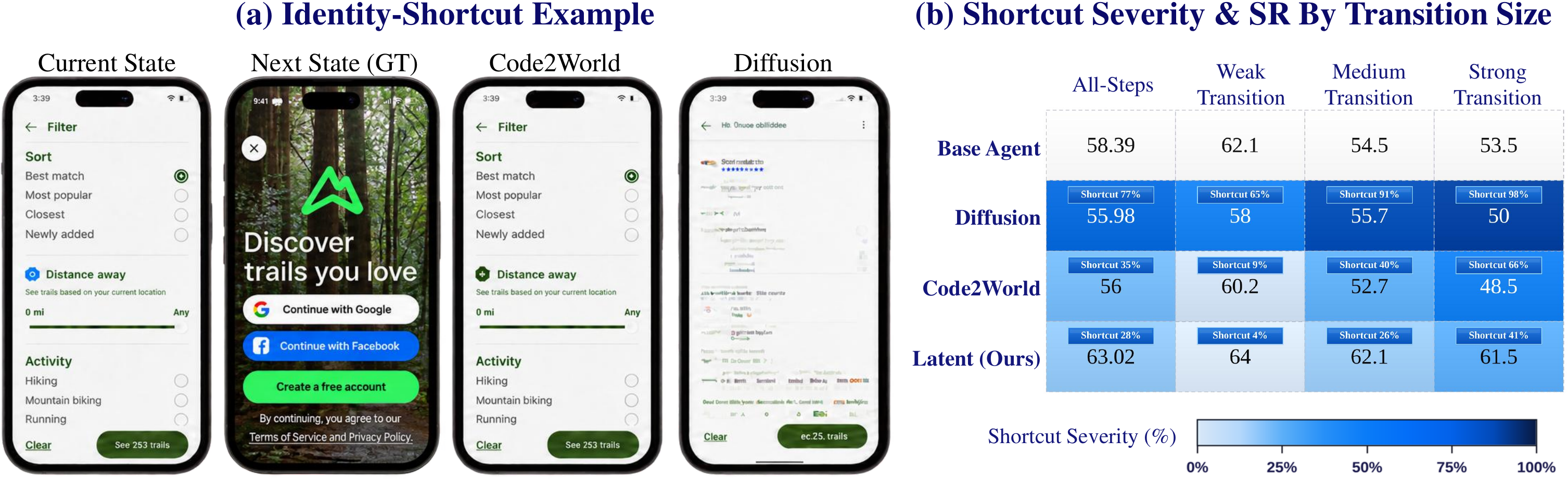}
    \vspace{-2mm}
    \caption{\textbf{Identity shortcut and SR across GUI transition magnitudes.} \textbf{(a)} Example of identity-preserving future prediction. \textbf{(b)} Downstream SR on a random 2.5K-sample AndroidControl subset. Shortcut severity is measured in each method's native prediction space---pixel space for baselines and latent space for ours---as required by each formulation. In practice, larger transitions coincide with higher shortcut severity and lower SR, while ours substantially attenuates these trends.}
    \vspace{-3mm}
    \label{fig:f4}
\end{figure}

\subsubsection{Identity Shortcut in GUI World Modeling}
However, this latent transition formulation alone does not explicitly ensure that the model captures the change induced by the action. That is, a point-wise objective independently pulls each prediction toward its paired next-state latent, without explicitly contrasting it against the current-state latent. Consequently, the model may preserve current-state information rather than capture the action-induced change, a behavior we refer to as the \textit{identity shortcut}. This issue is particularly relevant in GUI environments, where larger transitions can introduce substantial visual content that is weakly constrained by the current screen. Under such uncertainty, reusing current-state information can become an attractive shortcut, even under substantial visual changes, leading the model to preserve the existing interface rather than capture newly induced changes. We therefore examine whether this shortcut becomes more pronounced as the magnitude of GUI transitions increases.

As shown in Figure~\ref{fig:f4}(a), pixel-space world models using either renderable HTML prediction or diffusion-based screenshot generation can produce future screens visually close to the current GUI despite substantial changes in next state. Figure~\ref{fig:f4}(b) analyzes this behavior across transition magnitudes. We quantify transition magnitude by average absolute pixel change between grayscale $64\times64$ current and next screenshots, and group transitions into weak, medium, and strong regimes accordingly. We identify an identity shortcut when the predicted future remains close to the current screen by this pixel-change measure, despite a substantially larger current-to-next transition. Notably, the identity shortcut becomes increasingly pronounced as transition magnitude grows for pixel-space world models. Their downstream success rates also decrease for larger transitions, further motivating an objective that explicitly distinguishes the action-induced future state from the current state.

\subsubsection{Action-Aware Contrastive Learning for Latent GUI World Modeling}
We introduce an action-aware contrastive objective that encourages the predicted future representation to capture action-induced changes. Rather than supervising each prediction solely through point-wise matching to its target, we contrast it against alternative states that should remain distinguishable from the action-induced future. Specifically, we treat the predicted future latent $\hat{z}_{t+1}^{\,k}$ as the anchor and the next-state latent $z_{t+1}=\bar{\phi}(s_{t+1})$ as the positive. To directly discourage the identity shortcut, we encode the current screenshot as $z_t=\bar{\phi}(s_t)$ and use it as an explicit negative.

However, contrasting only against the current state provides limited discrimination among plausible action outcomes. For example, in the filter interface illustrated in Figure~\ref{fig:f4}(a), selecting ``Most popular'' or ``Closest'' produces similar next screens that differ mainly in the activated sorting option. Using only the current screen as a negative encourages either prediction to depart from the source state, but does not distinguish these action-specific outcomes from each other. We therefore maintain a first-in-first-out memory bank $\mathcal{B}$ of latents from other transitions, exposing each prediction to competing future states. These negatives encourage the predicted representation to encode the consequence specific to the selected action rather than merely differ from the current state.

To formulate the previously introduced contrastive objective, we $\ell_2$-normalize all latent representations and denote their normalized forms with a tilde. Let $\widetilde{\mathcal{B}}$ denote the normalized memory-bank entries. With the negative set $\mathcal{N}=\{\tilde{z}_t\}\cup\widetilde{\mathcal{B}}$ and temperature $\tau$, we define the transition objective:
\begin{equation}
    \mathcal{L}_{\mathrm{Transition}}
    =
    -\log
    \frac{
        \exp\!\big(
        \tilde{\hat{z}}_{t+1}^{\,k}\!\cdot\tilde{z}_{t+1}/\tau
        \big)
    }{
        \exp\!\big(
        \tilde{\hat{z}}_{t+1}^{\,k}\!\cdot\tilde{z}_{t+1}/\tau
        \big)
        +
        \sum_{n\in\mathcal{N}}
        \exp\!\big(
        \tilde{\hat{z}}_{t+1}^{\,k}\!\cdot n/\tau
        \big)
    }.
    \label{eq:transition_loss}
\end{equation}
This objective aligns predictions with the observed next state while separating them from the current state and competing futures. Thus, the transition objective discourages identity preservation and promotes action-specific transition information. Consistent with this objective, Figure~\ref{fig:f4}(b) shows that our latent model substantially reduces shortcut severity between $z_t$ and $\hat{z}_{t+1}$ directly in the shared latent space and maintains stable downstream SR across transition magnitudes. Finally, with $\lambda$ balancing action selection and predictive transition learning within the shared space, we jointly optimize the two objectives to couple action discrimination with predictive transition supervision:
\begin{equation}
    \mathcal{L}
    =
    \mathcal{L}_{\mathrm{Action}}
    +
    \lambda\,\mathcal{L}_{\mathrm{Transition}}.
\end{equation}

\section{Experiments}
\label{sec:exp}

\paragraph{Datasets and Evaluation Metrics.}
We evaluate our approach in both offline and online GUI interaction settings. For offline evaluation, we use AndroidControl~\citep{androidcontrol}, a large-scale dataset of Android interaction trajectories across diverse applications and tasks. We also use AITZ~\citep{aitz}, which provides mobile interaction trajectories with detailed action annotations. To assess out-of-distribution generalization, we further evaluate on GUI-Odyssey~\citep{lu2025guiodyssey}, a benchmark specifically designed for cross-app GUI navigation over longer interaction trajectories. For online evaluation, we use AndroidWorld~\citep{androidworld}, an interactive Android benchmark in which agents execute actions in real applications and are evaluated by task completion.

For offline evaluation, we report three step-level metrics. \textit{Type} measures whether the predicted action type matches the annotation. \textit{Grounding}, evaluated only on click actions, measures whether the predicted coordinate falls within a fixed radius ($14\%$ of the screen width) of the annotated point. \textit{SR} requires both the action type and argument to be correct. Specifically, a click must fall within the target element's bounding box, a text input must match the annotated content, and a scroll must match the annotated direction. For online evaluation, we report task success rate together with Tokens/Task, Mean Steps, and Run Time to measure both task performance and interaction efficiency.

\paragraph{Implementation Details.}
We evaluate our framework with GPT-5.6, Gemini 3.6 Flash, GUI-Owl-7B~\citep{guiowl}, Qwen3-VL-8B, and Qwen3-VL-32B~\citep{qwen3vl} as agent backbones. For world-model baselines, we compare against Code2World~\citep{code2world}, which predicts renderable code for future GUI states, and a diffusion-based world model~\citep{xu}, which generates future screenshots in pixel space. For comparison, each GUI agent provides the same candidate-action set to all world-model methods, which differ only in how these candidates are evaluated using predicted future information. All base agents follow the M3A~\citep{androidworld,xu} setup, receiving the screenshot and accessibility tree and executing an action at each interaction step. For our latent world-action model, we fine-tune Qwen3-VL-8B-Instruct using LoRA~\citep{lora} with rank $r{=}16$, scaling factor $\alpha{=}32$, and dropout $0.05$, applied to all attention and MLP projections. We optimize the model with AdamW using a learning rate of $4{\times}10^{-5}$ and an effective batch size of $8$ for $3$ epochs. We set the transition-loss weight to $\lambda{=}1.0$ and the contrastive temperature to $\tau{=}0.07$. Training takes $8$ hours on an NVIDIA A6000 GPU. Unless specified, the best and second-best results are highlighted in \textbf{bold} and \underline{underline}, respectively.



\begin{table*}[t]
    \centering
    \caption{Offline in-distribution (ID) evaluation on AndroidControl and AITZ across GUI agents.}
    \label{tab:offline}
    \resizebox{\textwidth}{!}{%
    \begin{tabular}{ll|ccc|ccc|cc}
        \toprule
        \textbf{Model} & \textbf{Settings}
        & \multicolumn{3}{c|}{\textbf{AndroidControl (ID)}}
        & \multicolumn{3}{c|}{\textbf{AITZ (ID)}}
        & \multicolumn{2}{c}{\textbf{Efficiency}} \\

        &
        & Type & Grounding & SR
        & Type & Grounding & SR
        & Tokens/Task & Run Time \\
        \midrule

        \multirow{4}{*}{GPT-5.6}
        & Base
        & \underline{71.92} & \textbf{60.46} & \textbf{59.84}
        & \underline{59.84} & \underline{47.99} & \underline{41.30}
        & \cellcolor{gray!20}\textbf{3585.57}
        & \cellcolor{gray!20}\textbf{1.14} \\

        & Code2World
        & 70.34 & \underline{59.61} & 58.22
        & 58.60 & 47.03 & 39.24
        & 12634.21 & 9.31 \\

        & Diffusion2Image
        & 68.31 & 55.01 & 57.94
        & 56.40 & 46.80 & 38.59
        & - & 135.20 \\

        & \cellcolor{cyan!10}\textbf{Latent (Ours)}
        & \cellcolor{cyan!10}\textbf{75.57}
        & \cellcolor{cyan!10}58.75
        & \cellcolor{cyan!10}\underline{58.52}
        & \cellcolor{cyan!10}\textbf{64.78}
        & \cellcolor{cyan!10}\textbf{48.21}
        & \cellcolor{cyan!10}\textbf{46.10}
        & \cellcolor{cyan!10}\underline{4260.90}
        & \cellcolor{cyan!10}\underline{1.46} \\[2pt]

        \midrule

        \multirow{4}{*}{Gemini-3.6-Flash}
        & Base
        & \underline{76.02} & \textbf{60.78} & \underline{57.25}
        & \underline{65.14} & \textbf{49.67} & \underline{45.19}
        & \cellcolor{gray!20}\textbf{2001.71}
        & \cellcolor{gray!20}\textbf{8.03} \\

        & Code2World
        & 75.80 & 57.61 & 55.50
        & 64.37 & \underline{48.08} & 44.15
        & 9557.18 & 21.20 \\

        & Diffusion2Image
        & 74.75 & 59.89 & 55.69
        & 64.08 & 47.61 & 43.51
        & - & 194.51 \\

        & \cellcolor{cyan!10}\textbf{Latent (Ours)}
        & \cellcolor{cyan!10}\textbf{80.36}
        & \cellcolor{cyan!10}\underline{60.50}
        & \cellcolor{cyan!10}\textbf{61.13}
        & \cellcolor{cyan!10}\textbf{66.19}
        & \cellcolor{cyan!10}45.80
        & \cellcolor{cyan!10}\textbf{47.25}
        & \cellcolor{cyan!10}\underline{2658.46}
        & \cellcolor{cyan!10}\underline{8.49} \\[2pt]





        \midrule
        \midrule

        \multirow{4}{*}{GUI-Owl-7B}
        & Base
        & \underline{73.23} & \underline{41.65} & \underline{40.40}
        & \underline{60.73} & \textbf{35.56} & \underline{34.78}
        & \cellcolor{gray!20}\textbf{1120.64}
        & \cellcolor{gray!20}\textbf{2.14} \\

        & Code2World
        & 68.52 & 36.08 & 31.80
        & 60.03 & 33.59 & 33.45
        & 3890.48 & 9.00 \\

        & Diffusion2Image
        & 66.39 & 37.72 & 33.11
        & 59.01 & 34.76 & 33.20
        & - & 100.88 \\

        & \cellcolor{cyan!10}\textbf{Latent (Ours)}
        & \cellcolor{cyan!10}\textbf{74.16}
        & \cellcolor{cyan!10}\textbf{43.55}
        & \cellcolor{cyan!10}\textbf{41.96}
        & \cellcolor{cyan!10}\textbf{60.99}
        & \cellcolor{cyan!10}\underline{35.23}
        & \cellcolor{cyan!10}\textbf{35.06}
        & \cellcolor{cyan!10}\underline{1181.68}
        & \cellcolor{cyan!10}\underline{2.22} \\[2pt]

        \midrule

        \multirow{4}{*}{Qwen3-VL-8B}
        & Base
        & \underline{75.94} & \underline{58.07} & \underline{55.56}
        & \underline{67.57} & 49.23 & \underline{44.52}
        & \cellcolor{gray!20}\textbf{1391.27}
        & \cellcolor{gray!20}\textbf{4.70} \\

        & Code2World
        & 75.21 & 56.80 & 54.04
        & 67.21 & 48.72 & 43.90
        & 8170.46 & 14.88 \\

        & Diffusion2Image
        & 74.29 & 56.73 & 54.96
        & 67.26 & \underline{49.30} & 43.31
        & - & 171.68 \\

        & \cellcolor{cyan!10}\textbf{Latent (Ours)}
        & \cellcolor{cyan!10}\textbf{77.15}
        & \cellcolor{cyan!10}\textbf{58.95}
        & \cellcolor{cyan!10}\textbf{57.09}
        & \cellcolor{cyan!10}\textbf{68.35}
        & \cellcolor{cyan!10}\textbf{50.40}
        & \cellcolor{cyan!10}\textbf{45.62}
        & \cellcolor{cyan!10}\underline{1702.17}
        & \cellcolor{cyan!10}\underline{4.83} \\[2pt]

        \bottomrule
    \end{tabular}%
    }
\end{table*}
\subsection{Offline Evaluation}
Table~\ref{tab:offline} summarizes the offline results on AndroidControl and AITZ. Our latent world-action model consistently improves action-type prediction across all evaluated backbones, with gains of $+0.9$ to $+4.3$ points on AndroidControl and $+0.3$ to $+4.9$ points on AITZ over the corresponding base agents. These improvements translate to higher step-level success rates in seven of eight backbone--dataset settings, with gains of up to $+4.8$ points. Improvements are more consistent for Type than for Grounding, as our rescoring mechanism changes which candidate action is selected but does not refine the grounding coordinates of individual candidates. Consequently, correcting the selected action type does not necessarily improve coordinate-level grounding performance across settings.

In contrast, pixel-space world-model baselines reduce SR relative to their base agents in every offline setting, with decreases of up to $8.6$ points. This pattern highlights the difficulty of using future-screen reconstruction as an intermediate step for action selection, whereas our approach incorporates future-state supervision into the latent representations used for candidate scoring. Compared with the base agents, our method increases Run Time by $0.08$--$0.46$ and Tokens/Task by $61$--$675$, while Code2World requires roughly $3.5$--$5.9\times$ more tokens and substantially longer inference. Also, Diffusion2Image incurs a larger Run Time due to iterative screenshot generation. Overall, latent predictive supervision consistently improves step-level action selection with modest inference overhead.

\begin{table*}[t]
    \centering
    \caption{Online evaluation on AndroidWorld across GUI agent backbones.}
    \label{tab:online}
    \resizebox{\textwidth}{!}{%
    \begin{tabular}{ll|cccc|ccc}
        \toprule
        \textbf{Model} & \textbf{Settings}
        & \multicolumn{4}{c|}{\textbf{AndroidWorld}}
        & \multicolumn{3}{c}{\textbf{Efficiency}} \\

        &
        & Easy SR
        & Medium SR
        & Hard SR
        & Overall SR
        & Tokens/Task
        & Mean Steps
        & Run Time \\
        \midrule

        \multirow{4}{*}{GPT-5.6}
        & Base
        & \underline{68.33} & 50.00 & 16.67 & 54.46
        & \cellcolor{gray!20}\underline{363.8K}
        & \cellcolor{gray!20}\underline{13.12}
        & \cellcolor{gray!20}\textbf{127.4} \\

        & Code2World
        & \underline{68.33} & \underline{52.94} & \underline{22.22} & \underline{56.25}
        & 525.4K & 13.71 & 370.9 \\

        & Diffusion2Image
        & \underline{68.33} & 47.06 & 16.67 & 53.57
        & - & 13.21 & 2320.3 \\

        & \cellcolor{cyan!10}\textbf{Latent (Ours)}
        & \cellcolor{cyan!10}\textbf{73.33}
        & \cellcolor{cyan!10}\textbf{58.82}
        & \cellcolor{cyan!10}\textbf{27.78}
        & \cellcolor{cyan!10}\textbf{61.61}
        & \cellcolor{cyan!10}\textbf{329.4K}
        & \cellcolor{cyan!10}\textbf{12.20}
        & \cellcolor{cyan!10}\underline{141.4} \\[2pt]

        \midrule

        \multirow{4}{*}{Gemini-3.6-Flash}
        & Base
        & 70.00 & \underline{52.94} & \underline{33.33} & \underline{58.93}
        & \cellcolor{gray!20}\underline{204.6K}
        & \cellcolor{gray!20}\underline{11.42}
        & \cellcolor{gray!20}\textbf{168.6} \\

        & Code2World
        & \underline{71.67} & \underline{52.94} & 22.22 & 58.04
        & 310.0K & 12.55 & 482.3 \\

        & Diffusion2Image
        & 68.33 & \underline{52.94} & 22.22 & 55.75
        & - & 11.68 & 2490.3 \\

        & \cellcolor{cyan!10}\textbf{Latent (Ours)}
        & \cellcolor{cyan!10}\textbf{73.33}
        & \cellcolor{cyan!10}\textbf{58.82}
        & \cellcolor{cyan!10}\textbf{38.89}
        & \cellcolor{cyan!10}\textbf{63.39}
        & \cellcolor{cyan!10}\textbf{200.4K}
        & \cellcolor{cyan!10}\textbf{11.12}
        & \cellcolor{cyan!10}\underline{228.1} \\[2pt]

        
        %
        %
        %

        \midrule
        \midrule

        \multirow{4}{*}{GUI-Owl-7B}
        & Base
        & 25.00 & \underline{5.88} & \underline{0.00} & 15.18
        & \cellcolor{gray!20}\textbf{226.1K}
        & \cellcolor{gray!20}\textbf{16.67}
        & \cellcolor{gray!20}\textbf{159.5} \\

        & Code2World
        & \underline{26.67} & \underline{5.88} & \textbf{5.56} & \underline{16.69}
        & 421.64K & 18.12 & 659.4 \\

        & Diffusion2Image
        & \underline{26.67} & \underline{5.88} & \underline{0.00} & 16.07
        & - & \underline{16.81} & 1260.2 \\

        & \cellcolor{cyan!10}\textbf{Latent (Ours)}
        & \cellcolor{cyan!10}\textbf{33.33}
        & \cellcolor{cyan!10}\textbf{11.76}
        & \cellcolor{cyan!10}\textbf{5.56}
        & \cellcolor{cyan!10}\textbf{22.32}
        & \cellcolor{cyan!10}\underline{229.0K}
        & \cellcolor{cyan!10}16.99
        & \cellcolor{cyan!10}\underline{205.3} \\[2pt]

        \midrule

        \multirow{4}{*}{Qwen3-VL-8B}
        & Base
        & 41.67 & 14.71 & 5.56 & 27.68
        & \cellcolor{gray!20}\textbf{105.9K}
        & \cellcolor{gray!20}\textbf{11.21}
        & \cellcolor{gray!20}\textbf{125.8} \\

        & Code2World
        & 38.33 & \underline{20.59} & \underline{11.11} & 28.57
        & 288.4K & 14.10 & 519.9 \\

        & Diffusion2Image
        & \underline{53.33} & 14.71 & 5.56 & \underline{33.93}
        & - & 13.59 & 789.9 \\

        & \cellcolor{cyan!10}\textbf{Latent (Ours)}
        & \cellcolor{cyan!10}\textbf{55.00}
        & \cellcolor{cyan!10}\textbf{26.47}
        & \cellcolor{cyan!10}\textbf{16.67}
        & \cellcolor{cyan!10}\textbf{40.18}
        & \cellcolor{cyan!10}\underline{129.3K}
        & \cellcolor{cyan!10}\underline{13.40}
        & \cellcolor{cyan!10}\underline{223.8} \\[2pt]

        \midrule

        \multirow{4}{*}{Qwen3-VL-32B}
        & Base
        & 56.67
        & 32.35
        & \underline{11.11}
        & 41.96
        & \cellcolor{gray!20}\textbf{320.8K}
        & \cellcolor{gray!20}\textbf{13.29}
        & \cellcolor{gray!20}\textbf{703.6} \\

        & Code2World
        & \underline{58.33}
        & \underline{35.29}
        & \underline{11.11}
        & \underline{43.75}
        & 597.3K
        & 16.84
        & 1453.6 \\
        
        & Diffusion2Image
        & 55.00
        & 26.47
        & 5.56
        & 38.39
        & -
        & 17.81
        & 3154.6 \\

        & \cellcolor{cyan!10}\textbf{Latent (Ours)}
        & \cellcolor{cyan!10}\textbf{60.00}
        & \cellcolor{cyan!10}\textbf{38.23}
        & \cellcolor{cyan!10}\textbf{16.67}
        & \cellcolor{cyan!10}\textbf{46.42}
        & \cellcolor{cyan!10}\underline{385.4K}
        & \cellcolor{cyan!10}\underline{16.26}
        & \cellcolor{cyan!10}\underline{960.0} \\[2pt]

        \bottomrule
    \end{tabular}%
    }
\end{table*}
\subsection{Online Evaluation}
Table~\ref{tab:online} presents the online evaluation results on the interactive AndroidWorld benchmark. Unlike offline step-level evaluation, AndroidWorld rolls out the agent's own actions, allowing improvements in action selection to influence subsequent states throughout the trajectory. Consistent with this intuition, our latent world-action model achieves the highest Overall SR across all five evaluated backbones, improving the corresponding base agents by $+4.5$ to $+12.5$ points. Notably, these gains persist across all task difficulty levels, with improvements of $+5.6$ to $+11.1$ points on Hard tasks. Complementing these online results, Figure~\ref{fig:f4}(b) shows that our latent model reduces shortcut severity relative to pixel-space world models while maintaining higher SR across transition magnitudes.

In contrast, pixel-space world models yield mixed changes in Overall SR while requiring substantially greater inference cost. Code2World provides only modest improvements across most backbones and requires approximately $2.1$--$4.1\times$ the Run Time of the base agents, while Diffusion2Image requires roughly $4.5$--$18.2\times$ longer execution. Our latent approach consistently achieves higher Overall SR than both pixel-space baselines while requiring substantially lower Run Time across all backbones. On the API-based backbones, GPT-5.6 and Gemini-3.6-Flash, our method further reduces both Mean Steps and total Tokens/Task relative to the base agents. Overall, our approach retains the benefits of world-model-based action evaluation while approaching the computational efficiency of base agents, bringing predictive GUI agents closer to practical real-time deployment.

\subsection{Ablation Studies}

\paragraph{Out-of-Distribution Generalization.}
Table~\ref{tab:ood} evaluates generalization to the out-of-distribution GUI-Odyssey benchmark. Our latent approach achieves the highest SR on both GUI-Owl-7B and Qwen3-VL-8B, improving over the corresponding base agents under the cross-app and longer-horizon distribution shift. In contrast, both pixel-space world models reduce SR below the base agents across the two backbones. These results suggest that latent predictive supervision transfers more reliably to unseen GUI environments, preserving the benefits of world-model-based action evaluation across distinct interaction distributions without explicit future-screen reconstruction.

\paragraph{Module Ablation.}
Table~\ref{tab:ablation} isolates each component's contribution on AndroidWorld. Group~I corresponds to the base agent without proposed modules. Adding the action head (Group~II) improves SR from $27.68$ to $36.36$, establishing a latent candidate-rescoring baseline before world-model supervision. Group~III incorporates the transition head to learn future-state information. Without contrastive learning, we train it with point-wise cosine regression, $\mathcal{L}_{\mathrm{GroupIII}}=1-\cos(\hat{z}_{t+1}^{\,k},z_{t+1})$, between predicted and ground-truth next-state latents. This predictive supervision raises SR by $+1.82$ points to $38.18$ over Group~II, demonstrating additional benefit beyond action rescoring. Finally, Group~IV replaces point-wise matching with our action-aware contrastive objective, raising SR by another $+2.00$ points to $40.18$. This gain isolates the benefit of explicitly distinguishing the action-induced future from the current state and competing future states, supporting our contrastive formulation. Overall, the model improves SR by $+12.50$ points over the base agent. Importantly, these components operate in latent space, preserving efficiency without future-screen generation.

\begin{table*}[t]
    \centering

    \begin{minipage}[t]{0.53\textwidth}
        \centering

        \parbox[t][1.7\baselineskip][t]{\linewidth}{
            \centering
            \caption{OOD evaluation on GUI-Odyssey.}
            \label{tab:ood}
        }

        \begin{adjustbox}{max width=\linewidth}
        \begin{tabular}{ll|ccc}
            \toprule
            \textbf{Model} & \textbf{Settings}
            & \multicolumn{3}{c}{\textbf{GUI-Odyssey (OOD)}} \\
            & & Type & Grounding & SR \\
            \midrule

            \multirow{4}{*}{GUI-Owl-7B}
            & Base
            & \underline{69.09}
            & \underline{38.97}
            & \underline{33.73} \\

            & Code2World
            & 66.77 & 32.13 & 27.78 \\

            & Diffusion2Image
            & 67.57 & 31.11 & 26.83 \\

            & \cellcolor{cyan!10}\textbf{Latent (Ours)}
            & \cellcolor{cyan!10}\textbf{70.15}
            & \cellcolor{cyan!10}\textbf{39.07}
            & \cellcolor{cyan!10}\textbf{34.73} \\[2pt]

            \midrule

            \multirow{4}{*}{Qwen3-VL-8B}
            & Base
            & 77.72
            & \underline{62.45}
            & \underline{52.07} \\

            & Code2World
            & 77.28 & 59.70 & 49.94 \\

            & Diffusion2Image
            & \textbf{77.93}
            & \textbf{62.98}
            & 51.44 \\

            & \cellcolor{cyan!10}\textbf{Latent (Ours)}
            & \cellcolor{cyan!10}\underline{77.91}
            & \cellcolor{cyan!10}62.26
            & \cellcolor{cyan!10}\textbf{52.13} \\[2pt]

            \bottomrule
        \end{tabular}
        \end{adjustbox}

    \end{minipage}%
    \hfill
    \begin{minipage}[t]{0.45\textwidth}
        \centering

        \parbox[t][3.7\baselineskip][t]{\linewidth}{
            \centering
            \caption{Module ablation of the proposed method on AndroidWorld. Checkmarks indicate the inclusion of the components.}
            \label{tab:ablation}
        }

        \begin{adjustbox}{max width=\linewidth}
        \begin{tabular}{@{}ccccc@{}}
            \toprule
            \textbf{Group}
            & \multicolumn{3}{c}{\textbf{Components}}
            & \textbf{SR} \\
            \cmidrule(lr){2-4}

            & \shortstack{Action\\Head}
            & \shortstack{Contrastive\\Learning}
            & \shortstack{Transition\\Head}
            & \\

            \midrule

            I
            & \xmark
            & \xmark
            & \xmark
            & 27.68 \\

            II
            & \checkmark
            & \xmark
            & \xmark
            & 36.36 \\

            III
            & \checkmark
            & \xmark
            & \checkmark
            & 38.18 \\

            \rowcolor{cyan!10}
            IV
            & \checkmark
            & \checkmark
            & \checkmark
            & \textbf{40.18} \\

            \bottomrule
        \end{tabular}
        \end{adjustbox}

    \end{minipage}

\end{table*}
\begin{figure}[t]
    \centering
    \includegraphics[width=\linewidth]{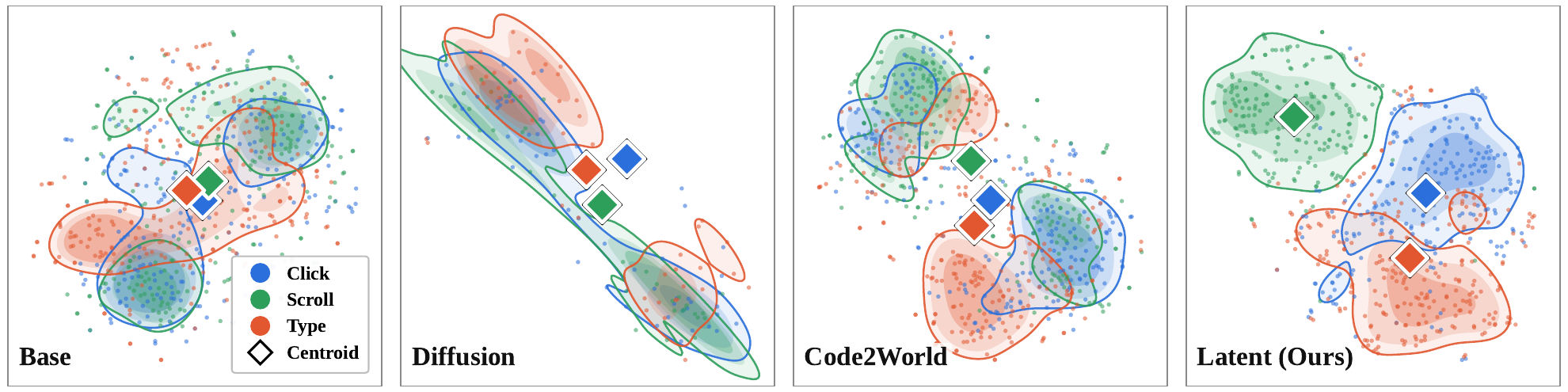}
    \vspace{-6mm}
    \caption{\textbf{t-SNE visualization of action-induced future representations.} Our model better aligns its predicted futures with action-specific clusters than the base agent and pixel-space world models.}
    \label{fig:f5}
    \vspace{-3mm}
\end{figure}

\paragraph{Qualitative Analysis of Action-Specific Future Representations.}
We qualitatively examine future representations from AndroidControl to assess whether they capture action-specific transition information useful for action selection beyond current screen. Given the identity-preserving tendency observed earlier, we seek to disentangle action-induced structure from current-state information. A visualization can be misleading because similar GUI states often admit similar action types, such that apparent action-specific clustering may arise from screen--action correlations rather than modeled action-induced transitions. To reduce this confounding, we cluster current states by their latent representations $z_t$ using $k$-means. For each sample, we subtract the mean predicted representation of its current-state cluster and apply t-SNE to the residual. This procedure suppresses information shared within similar current-state contexts, allowing us to examine whether the remaining variation is structured by the executed action. As shown in Figure~\ref{fig:f5}, the Base and Diffusion representations remain mixed across action types, while Code2World exhibits only partial separation. In contrast, our latent model forms the clearest action-specific structure around the centroids. This result provides qualitative evidence that our predictive representations capture how different actions induce distinct future-state changes beyond correlations attributable to the current screen.

\section{Conclusion}
\label{sec:conclusion}
We revisit GUI world modeling by asking which future information matters for action selection. Rather than reconstructing future screens, we introduce a latent world-action formulation that jointly learns action evaluation and predictive transitions with action-aware contrastive learning in a shared space, capturing action-induced future information without costly screen generation and re-perception. Empirically, we observe consistent performance gains over base agents and pixel-space world models, with substantially improved inference speed over pixel-space world models. These results suggest a direction for GUI world modeling: future prediction need not reproduce what the future looks like, but should instead capture the decision-relevant consequences of agent actions.

\bibliography{iclr2027_conference}

@inproceedings{cogagent,
  author       = {Wenyi Hong and
                  Weihan Wang and
                  Qingsong Lv and
                  Jiazheng Xu and
                  Wenmeng Yu and
                  Junhui Ji and
                  Yan Wang and
                  Zihan Wang and
                  Yuxiao Dong and
                  Ming Ding and
                  Jie Tang},
  title        = {CogAgent: {A} Visual Language Model for {GUI} Agents},
  booktitle    = {{IEEE/CVF} Conference on Computer Vision and Pattern Recognition,
                  {CVPR} 2024, Seattle, WA, USA, June 16-22, 2024},
  publisher    = {{IEEE}},
  year         = {2024},
}

@inproceedings{tora,
  author       = {Zhibin Gou and
                  Zhihong Shao and
                  Yeyun Gong and
                  Yelong Shen and
                  Yujiu Yang and
                  Minlie Huang and
                  Nan Duan and
                  Weizhu Chen},
  title        = {ToRA: {A} Tool-Integrated Reasoning Agent for Mathematical Problem
                  Solving},
  booktitle    = {The Twelfth International Conference on Learning Representations,
                  {ICLR} 2024, Vienna, Austria, May 7-11, 2024},
  publisher    = {OpenReview.net},
  year         = {2024},
}

@article{scope,
  author       = {Zizhao Tong and
                  Yeying Jin and
                  Hongfeng Lai and
                  Zeqing Wang and
                  Zhaohu Xing and
                  Kexu Cheng and
                  Haoran Xu and
                  Zhao Pu and
                  Shangwen Zhu and
                  Ruili Feng and
                  Jian Zhao and
                  Yan Zhang and
                  Hao Tang and
                  Ling Shao},
  title        = {{SCOPE:} Simulating Cross-game Operations in Playable Environments
                  for {FPS} World Models},
  journal      = {CoRR},
  volume       = {abs/2605.23345},
  year         = {2026},
}

@inproceedings{raft,
  author       = {Zachary Teed and
                  Jia Deng},
  title        = {{RAFT:} Recurrent All-Pairs Field Transforms for Optical Flow},
  booktitle    = {Computer Vision - {ECCV} 2020 - 16th European Conference, Glasgow,
                  UK, August 23-28, 2020, Proceedings, Part {II}},
  volume       = {12347},
  pages        = {402--419},
  publisher    = {Springer},
  year         = {2020},
}

@inproceedings{tvqa,
  author       = {Jie Lei and
                  Licheng Yu and
                  Mohit Bansal and
                  Tamara L. Berg},
  title        = {{TVQA:} Localized, Compositional Video Question Answering},
  booktitle    = {Proceedings of the 2018 Conference on Empirical Methods in Natural
                  Language Processing, Brussels, Belgium, October 31 - November 4, 2018},
  pages        = {1369--1379},
  publisher    = {Association for Computational Linguistics},
  year         = {2018},
}

@article{mobileguirl,
  author       = {Yucheng Shi and
                  Wenhao Yu and
                  Zaitang Li and
                  Yonglin Wang and
                  Hongming Zhang and
                  Ninghao Liu and
                  Haitao Mi and
                  Dong Yu},
  title        = {MobileGUI-RL: Advancing Mobile {GUI} Agent through Reinforcement Learning
                  in Online Environment},
  journal      = {CoRR},
  volume       = {abs/2507.05720},
  year         = {2025},
}

@article{arpo,
  author       = {Fanbin Lu and
                  Zhisheng Zhong and
                  Shu Liu and
                  Chi{-}Wing Fu and
                  Jiaya Jia},
  title        = {ARPO:End-to-End Policy Optimization for {GUI} Agents with Experience
                  Replay},
  journal      = {CoRR},
  volume       = {abs/2505.16282},
  year         = {2025},
}

@inproceedings{guireflection,
  author       = {Penghao Wu and
                  Shengnan Ma and
                  Bo Wang and
                  Jiaheng Yu and
                  Lewei Lu and
                  Ziwei Liu},
  title        = {GUI-Reflection: Empowering Multimodal {GUI} Models with Self-Reflection
                  Behavior},
  booktitle    = {Advances in Neural Information Processing Systems 38: Annual Conference
                  on Neural Information Processing Systems 2025, NeurIPS 2025, San Diego,
                  CA, USA, December 2-7, 2025 / Mexico City, Mexico, November 30 - December
                  5, 2025},
  year         = {2025},
}

@inproceedings{guirise,
  author       = {Tao Liu and
                  Chongyu Wang and
                  Rongjie Li and
                  Yingchen Yu and
                  Xuming He and
                  Song Bai},
  title        = {GUI-Rise: Structured Reasoning and History Summarization for {GUI}
                  Navigation},
  booktitle    = {Advances in Neural Information Processing Systems 38: Annual Conference
                  on Neural Information Processing Systems 2025, NeurIPS 2025, San Diego,
                  CA, USA, December 2-7, 2025 / Mexico City, Mexico, November 30 - December
                  5, 2025},
  year         = {2025},
}

@article{mobileagent,
  author       = {Junyang Wang and
                  Haiyang Xu and
                  Jiabo Ye and
                  Ming Yan and
                  Weizhou Shen and
                  Ji Zhang and
                  Fei Huang and
                  Jitao Sang},
  title        = {Mobile-Agent: Autonomous Multi-Modal Mobile Device Agent with Visual
                  Perception},
  journal      = {CoRR},
  year         = {2024},
}

@inproceedings{uivision,
  author       = {Shravan Nayak and
                  Xiangru Jian and
                  Kevin Qinghong Lin and
                  Juan A. Rodr{\'{\i}}guez and
                  Montek Kalsi and
                  Nicolas Chapados and
                  M. Tamer {\"{O}}zsu and
                  Aishwarya Agrawal and
                  David V{\'{a}}zquez and
                  Christopher Pal and
                  Perouz Taslakian and
                  Spandana Gella and
                  Sai Rajeswar},
  title        = {UI-Vision: {A} Desktop-centric {GUI} Benchmark for Visual Perception
                  and Interaction},
  booktitle    = {Forty-second International Conference on Machine Learning, {ICML}
                  2025, Vancouver, BC, Canada, July 13-19, 2025},
  volume       = {267},
  year         = {2025},
}

@inproceedings{showui,
  author       = {Kevin Qinghong Lin and
                  Linjie Li and
                  Difei Gao and
                  Zhengyuan Yang and
                  Shiwei Wu and
                  Zechen Bai and
                  Stan Weixian Lei and
                  Lijuan Wang and
                  Mike Zheng Shou},
  title        = {ShowUI: One Vision-Language-Action Model for {GUI} Visual Agent},
  booktitle    = {{IEEE/CVF} Conference on Computer Vision and Pattern Recognition,
                  {CVPR} 2025, Nashville, TN, USA, June 11-15, 2025},
  pages        = {19498--19508},
  year         = {2025},
}

@inproceedings{drivelaw,
    author    = {Xia, Tianze and Li, Yongkang and Zhou, Lijun and Yao, Jingfeng and Xiong, Kaixin and Sun, Haiyang and Wang, Bing and Ma, Kun and Chen, Guang and Ye, Hangjun and Liu, Wenyu and Wang, Xinggang},
    title     = {DriveLaW: Unifying Planning and Video Generation in a Latent Driving World},
    booktitle = {Proceedings of the IEEE/CVF Conference on Computer Vision and Pattern Recognition (CVPR)},
    year      = {2026},
    pages     = {39701-39712}
}

@inproceedings{latentcot,
    author    = {Tan, Shuhan and Chitta, Kashyap and Chen, Yuxiao and Tian, Ran and You, Yurong and Wang, Yan and Luo, Wenjie and Cao, Yulong and Kr\"ahenb\"uhl, Philipp and Pavone, Marco and Ivanovic, Boris},
    title     = {Latent Chain-of-Thought World Modeling for End-to-End Autonomous Driving},
    booktitle = {Proceedings of the IEEE/CVF Conference on Computer Vision and Pattern Recognition (CVPR)},
    year      = {2026},
    pages     = {39724-39733}
}

@inproceedings{driveworldvla,
    title={DriveWorld-{VLA}: Unified Latent-Space World Modeling with Vision{\textendash}Language{\textendash}Action for Autonomous Driving},
    author={Feiyang Jia and Lin Liu and Ziying Song and Caiyan Jia and Hangjun Ye and Xiaoshuai Hao and Long Chen},
    booktitle={Forty-third International Conference on Machine Learning},
    year={2026},
}

@inproceedings{lora,
  author       = {Edward J. Hu and
                  Yelong Shen and
                  Phillip Wallis and
                  Zeyuan Allen{-}Zhu and
                  Yuanzhi Li and
                  Shean Wang and
                  Lu Wang and
                  Weizhu Chen},
  title        = {LoRA: Low-Rank Adaptation of Large Language Models},
  booktitle    = {The Tenth International Conference on Learning Representations, {ICLR}
                  2022, Virtual Event, April 25-29, 2022},
  year         = {2022},
}

@article{univla,
  author       = {Qingwen Bu and
                  Yanting Yang and
                  Jisong Cai and
                  Shenyuan Gao and
                  Guanghui Ren and
                  Maoqing Yao and
                  Ping Luo and
                  Hongyang Li},
  title        = {UniVLA: Learning to Act Anywhere with Task-centric Latent Actions},
  journal      = {CoRR},
  volume       = {abs/2505.06111},
  year         = {2025},
}

@inproceedings{zhou,
  author       = {Luowei Zhou and
                  Chenliang Xu and
                  Jason J. Corso},
  title        = {Towards Automatic Learning of Procedures From Web Instructional Videos},
  booktitle    = {Proceedings of the Thirty-Second {AAAI} Conference on Artificial Intelligence,
                  (AAAI-18), the 30th innovative Applications of Artificial Intelligence
                  (IAAI-18), and the 8th {AAAI} Symposium on Educational Advances in
                  Artificial Intelligence (EAAI-18), New Orleans, Louisiana, USA, February
                  2-7, 2018},
  pages        = {7590--7598},
  publisher    = {{AAAI} Press},
  year         = {2018},
}

@article{qwen3vl,
  author       = {Qwen Team},
  title        = {Qwen3-VL Technical Report},
  journal      = {CoRR},
  volume       = {abs/2511.21631},
  year         = {2025},
}

@article{uitars,
  author       = {Yujia Qin and
                  Yining Ye and
                  Junjie Fang and
                  Haoming Wang and
                  Shihao Liang and
                  Shizuo Tian and
                  Junda Zhang and
                  Jiahao Li and
                  Yunxin Li and
                  Shijue Huang and
                  Wanjun Zhong and
                  Kuanye Li and
                  Jiale Yang and
                  Yu Miao and
                  Woyu Lin and
                  Longxiang Liu and
                  Xu Jiang and
                  Qianli Ma and
                  Jingyu Li and
                  Xiaojun Xiao and
                  Kai Cai and
                  Chuang Li and
                  Yaowei Zheng and
                  Chaolin Jin and
                  Chen Li and
                  Xiao Zhou and
                  Minchao Wang and
                  Haoli Chen and
                  Zhaojian Li and
                  Haihua Yang and
                  Haifeng Liu and
                  Feng Lin and
                  Tao Peng and
                  Xin Liu and
                  Guang Shi},
  title        = {{UI-TARS:} Pioneering Automated {GUI} Interaction with Native Agents},
  journal      = {CoRR},
  volume       = {abs/2501.12326},
  year         = {2025},
}

@inproceedings{androidinthewild,
  author       = {Christopher Rawles and
                  Alice Li and
                  Daniel Rodriguez and
                  Oriana Riva and
                  Timothy P. Lillicrap},
  title        = {AndroidInTheWild: {A} Large-Scale Dataset For Android Device Control},
  booktitle    = {Advances in Neural Information Processing Systems 36: Annual Conference
                  on Neural Information Processing Systems 2023, NeurIPS 2023, New Orleans,
                  LA, USA, December 10 - 16, 2023},
  year         = {2023},
}

@inproceedings{syed,
  author       = {Syed Fatiul Huq and
                  Ziyao He and
                  Yirui He and
                  Sam Malek},
  title        = {Bridging the Gap between Automated Intervention and Actual User Experience:
                  {A} Mixed-Methods Study on Mobile Accessibility Issues for Screen
                  Reader Users},
  booktitle    = {Proceedings of the 2026 {CHI} Conference on Human Factors in Computing
                  Systems, {CHI} 2026, Barcelona, Spain, April 13-17, 2026},
  publisher    = {{ACM}},
  year         = {2026},
}

@inproceedings{appagent,
  author       = {Chi Zhang and
                  Zhao Yang and
                  Jiaxuan Liu and
                  Yanda Li and
                  Yucheng Han and
                  Xin Chen and
                  Zebiao Huang and
                  Bin Fu and
                  Gang Yu},
  title        = {AppAgent: Multimodal Agents as Smartphone Users},
  booktitle    = {Proceedings of the 2025 {CHI} Conference on Human Factors in Computing
                  Systems, {CHI} 2025, YokohamaJapan, 26 April 2025- 1 May 2025},
  publisher    = {{ACM}},
  year         = {2025},
}

@inproceedings{screen,
  author       = {Xiaoyi Zhang and
                  Lilian de Greef and
                  Amanda Swearngin and
                  Samuel White and
                  Kyle I. Murray and
                  Lisa Yu and
                  Qi Shan and
                  Jeffrey Nichols and
                  Jason Wu and
                  Chris Fleizach and
                  Aaron Everitt and
                  Jeffrey P. Bigham},
  title        = {Screen Recognition: Creating Accessibility Metadata for Mobile Applications
                  from Pixels},
  booktitle    = {{CHI} '21: {CHI} Conference on Human Factors in Computing Systems,
                  Virtual Event / Yokohama, Japan, May 8-13, 2021},
  publisher    = {{ACM}},
  year         = {2021},
}

@inproceedings{mobilegpt,
  author       = {Sunjae Lee and
                  Junyoung Choi and
                  Jungjae Lee and
                  Munim Hasan Wasi and
                  Hojun Choi and
                  Steven Y. Ko and
                  Sangeun Oh and
                  Insik Shin},
  title        = {MobileGPT: Augmenting {LLM} with Human-like App Memory for Mobile
                  Task Automation},
  booktitle    = {Proceedings of the 30th Annual International Conference on Mobile
                  Computing and Networking, {ACM} MobiCom 2024, Washington D.C., DC,
                  USA, November 18-22, 2024},
  publisher    = {{ACM}},
  year         = {2024},
}

@article{lawam,
  author       = {Jialei Chen and
                  Kai Wang and
                  Kang Chen and
                  Shuaihang Chen and
                  Feng Gao and
                  Wenhao Tang and
                  Zhiyuan Li and
                  Weilin Liu and
                  Zhuyu Yao and
                  Boxun Li and
                  Yuanbo Xu and
                  Chao Yu},
  title        = {LaWAM: Latent World Action Models for Efficient Dynamics-Aware Robot
                  Policies},
  journal      = {CoRR},
  volume       = {abs/2606.15768},
  year         = {2026},
}

@article{guiowl,
  author       = {Jiabo Ye and
                  Xi Zhang and
                  Haiyang Xu and
                  Haowei Liu and
                  Junyang Wang and
                  Zhaoqing Zhu and
                  Ziwei Zheng and
                  Feiyu Gao and
                  Junjie Cao and
                  Zhengxi Lu and
                  Jitong Liao and
                  Qi Zheng and
                  Fei Huang and
                  Jingren Zhou and
                  Ming Yan},
  title        = {Mobile-Agent-v3: Fundamental Agents for {GUI} Automation},
  journal      = {CoRR},
  volume       = {abs/2508.15144},
  year         = {2025},
}

@article{code2world,
  author       = {Yuhao Zheng and
                  Li'an Zhong and
                  Yi Wang and
                  Rui Dai and
                  Kaikui Liu and
                  Xiangxiang Chu and
                  Linyuan Lv and
                  Philip Torr and
                  Kevin Qinghong Lin},
  title        = {Code2World: {A} {GUI} World Model via Renderable Code Generation},
  journal      = {CoRR},
  volume       = {abs/2602.09856},
  year         = {2026},
}

@inproceedings{aitz,
  author       = {Jiwen Zhang and
                  Jihao Wu and
                  Yihua Teng and
                  Minghui Liao and
                  Nuo Xu and
                  Xiao Xiao and
                  Zhongyu Wei and
                  Duyu Tang},
  title        = {Android in the Zoo: Chain-of-Action-Thought for {GUI} Agents},
  booktitle    = {Findings of the Association for Computational Linguistics: {EMNLP}
                  2024, Miami, Florida, USA, November 12-16, 2024},
  volume       = {{EMNLP} 2024},
  pages        = {12016--12031},
  year         = {2024},
}

@article{gu,
  author       = {Yu Gu and
                  Kai Zhang and
                  Yuting Ning and
                  Boyuan Zheng and
                  Boyu Gou and
                  Tianci Xue and
                  Cheng Chang and
                  Sanjari Srivastava and
                  Yanan Xie and
                  Peng Qi and
                  Huan Sun and
                  Yu Su},
  title        = {Is Your {LLM} Secretly a World Model of the Internet? Model-Based
                  Planning for Web Agents},
  journal      = {Trans. Mach. Learn. Res.},
  volume       = {2025},
  year         = {2025},
}

@article{mobiledreamer,
  author       = {Yilin Cao and
                  Yufeng Zhong and
                  Zhixiong Zeng and
                  Liming Zheng and
                  Jing Huang and
                  Haibo Qiu and
                  Peng Shi and
                  Wenji Mao and
                  Wan Guanglu},
  title        = {MobileDreamer: Generative Sketch World Model for {GUI} Agent},
  journal      = {CoRR},
  volume       = {abs/2601.04035},
  year         = {2026},
}

@inproceedings{chae,
title={Web Agents with World Models: Learning and Leveraging Environment Dynamics in Web Navigation},
author={Hyungjoo Chae and Namyoung Kim and Kai Tzu-iunn Ong and Minju Gwak and Gwanwoo Song and Jihoon Kim and Sunghwan Kim and Dongha Lee and Jinyoung Yeo},
booktitle={The Thirteenth International Conference on Learning Representations},
year={2025},
}

@misc{xu,
      title={How Mobile World Model Guides GUI Agents?}, 
      author={Weikai Xu and Kun Huang and Yunren Feng and Jiaxing Li and Yuhan Chen and Yuxuan Liu and Zhizheng Jiang and Heng Qu and Pengzhi Gao and Wei Liu and Jian Luan and Xiaolin Hu and Bo An},
      year={2026},
      eprint={2605.10347},
      archivePrefix={arXiv},
      primaryClass={cs.AI},
      url={https://arxiv.org/abs/2605.10347}, 
}

@inproceedings{vimo,
    title={ViMo: A Generative Visual {GUI} World Model for App Agents},
    author={Dezhao Luo and Bohan Tang and Kang Li and Georgios Papoudakis and Jifei Song and Shaogang Gong and Jianye HAO and Jun Wang and Kun Shao},
    booktitle={The Fourteenth International Conference on Learning Representations},
    year={2026},
}

@inproceedings{reds,
  author       = {Seungjun Nah and
                  Sungyong Baik and
                  Seokil Hong and
                  Gyeongsik Moon and
                  Sanghyun Son and
                  Radu Timofte and
                  Kyoung Mu Lee},
  title        = {{NTIRE} 2019 Challenge on Video Deblurring and Super-Resolution: Dataset
                  and Study},
  booktitle    = {{IEEE} Conference on Computer Vision and Pattern Recognition Workshops,
                  {CVPR} Workshops 2019, Long Beach, CA, USA, June 16-20, 2019},
  pages        = {1996--2005},
  publisher    = {Computer Vision Foundation / {IEEE}},
  year         = {2019},
}

@inproceedings{androidcontrol,
  author       = {Wei Li and
                  William E. Bishop and
                  Alice Li and
                  Christopher Rawles and
                  Folawiyo Campbell{-}Ajala and
                  Divya Tyamagundlu and
                  Oriana Riva},
  title        = {On the Effects of Data Scale on {UI} Control Agents},
  booktitle    = {Advances in Neural Information Processing Systems 37: Annual Conference
                  on Neural Information Processing Systems 2024, NeurIPS 2024, Vancouver,
                  BC, Canada, December 10 - 15, 2024},
  year         = {2024},
}

@inproceedings{androidworld,
  author       = {Christopher Rawles and
                  Sarah Clinckemaillie and
                  Yifan Chang and
                  Jonathan Waltz and
                  Gabrielle Lau and
                  Marybeth Fair and
                  Alice Li and
                  William E. Bishop and
                  Wei Li and
                  Folawiyo Campbell{-}Ajala and
                  Daniel Kenji Toyama and
                  Robert James Berry and
                  Divya Tyamagundlu and
                  Timothy P. Lillicrap and
                  Oriana Riva},
  title        = {AndroidWorld: {A} Dynamic Benchmarking Environment for Autonomous
                  Agents},
  booktitle    = {The Thirteenth International Conference on Learning Representations,
                  {ICLR} 2025, Singapore, April 24-28, 2025},
  publisher    = {OpenReview.net},
  year         = {2025},
}

@inproceedings{lu2025guiodyssey,
  title={Guiodyssey: A comprehensive dataset for cross-app gui navigation on mobile devices},
  author={Lu, Quanfeng and Shao, Wenqi and Liu, Zitao and Du, Lingxiao and Meng, Fanqing and Li, Boxuan and Chen, Botong and Huang, Siyuan and Zhang, Kaipeng and Luo, Ping},
  booktitle={2025 IEEE/CVF International Conference on Computer Vision (ICCV)},
  pages={22404--22414},
  year={2025},
  organization={IEEE}
}
\bibliographystyle{iclr2027_conference}

%
\appendix
\section*{Appendix}

\setcounter{figure}{0}
\setcounter{table}{0}
\setcounter{equation}{0}

\renewcommand{\thefigure}{A\arabic{figure}}
\renewcommand{\thetable}{A\arabic{table}}
\renewcommand{\theequation}{A\arabic{equation}}

\ifdefined\theHfigure
  \renewcommand{\theHfigure}{appendix.\arabic{figure}}
\fi
\ifdefined\theHtable
  \renewcommand{\theHtable}{appendix.\arabic{table}}
\fi
\ifdefined\theHequation
  \renewcommand{\theHequation}{appendix.\arabic{equation}}
\fi


\startcontents[supplement]

\begingroup
  \color{appendixblue}

  \ifdefined\hypersetup
    \hypersetup{
      colorlinks=true,
      linkcolor=appendixblue
    }
  \fi

  \section*{\textcolor{black}{Contents}}

  \printcontents[supplement]{}{1}{%
      \setcounter{tocdepth}{2}%
    }
\endgroup

%


\section{Limitations and Future Work}

\paragraph{Dependence on the Base Agent's Candidates.}
Our latent world-action model re-ranks candidate actions proposed by the frozen base GUI agent and does not generate new actions. When the correct action is absent from the candidate set, the model cannot recover it. Its performance is therefore bounded by the quality and coverage of the base agent's proposals. Expanding candidate coverage through more diverse sampling, adaptive candidate-set sizes, or improved candidate generation is a direction for reducing this dependence, when candidate proposals are narrow or redundant.

\paragraph{Selective Invocation.}
Our model is invoked only when the confidence-based gate identifies a low-confidence decision. Consequently, errors made by the base agent with high confidence can bypass latent world-model evaluation, and terminal decisions such as completing or abandoning a task are not rescored in our current protocol. Replacing the fixed confidence threshold with a learned, uncertainty-aware invocation policy could improve coverage while retaining the computational benefit of selective rescoring and accounting for the expected value of additional evaluation.

\paragraph{Single-Step Latent Prediction.}
The transition objective models only the immediate action-conditioned next state during training, and the transition head is discarded at inference. This design enables efficient single-pass action scoring, but it does not explicitly reason over longer-horizon consequences or provide interpretable reconstructions of predicted future states. Extending the objective to multi-step latent rollouts could support longer-term action evaluation, while lightweight decoders or probing modules could improve interpretability without requiring full-screen generation.

\paragraph{Future Work.}
These limitations suggest several complementary extensions. Future work can combine latent action evaluation with stronger candidate generation, replace fixed confidence gating with learned invocation policies, and extend one-step transition learning to multi-step latent dynamics for longer-horizon decision-making. Finally, training with candidates from multiple base agents and evaluating across web and desktop interfaces would help assess whether latent action-conditioned transition modeling generalizes beyond the mobile GUI environments considered here.

\section{Experimental Details}
This section provides additional details on the experimental setup, including implementation, datasets, evaluation protocols, and analysis settings.

\subsection{Device Information}
All experiments were conducted on Ubuntu 22.04 LTS using eight NVIDIA RTX A6000 GPUs and dual AMD EPYC 7413 CPUs. Training used approximately 31 GiB of GPU memory per device, while evaluation used approximately 22 GiB for the judge and 26 GiB for the Qwen3-VL-8B agent. We used Python 3.11.15, PyTorch 2.6.0 with CUDA 12.4, and a random seed of 3 for training and 30 for AndroidWorld evaluation.

\subsection{Implementation Details}
This section provides additional implementation details of the baseline protocols, model architecture, training and inference procedure, hyperparameters, and analysis settings.

\subsubsection{World-Model Baseline Protocol for GUI Agents}
In this section, we describe how GUI world models support action selection through the ``Render-and-Re-Perceive'' pipeline~\citep{code2world}. This common protocol consists of two stages: (1) explicit future-state prediction by a world model and (2) world-model-guided GUI action selection.

\paragraph{Stage 1: Future-State GUI World Modeling.}
Recent works on GUI agents have increasingly explored future-state world modeling, enabling more robust action decision-making by anticipating the consequences of candidate actions. Given the current GUI state $s_t$, a user-specified goal $g$, and interaction history $h_t$, a GUI agent $\mathcal{G}$ first proposes $K$ candidate actions, after which a world model $\mathcal{W}$ predicts the corresponding future state $\hat{s}_{t+1}^k$ for each candidate as follows:
\begin{equation}
    \boldsymbol{\mathcal{A}}_t = \mathcal{G}(s_t,g,h_t)=\{a_t^k\}_{k=1}^{K},
    \qquad
    \hat{s}_{t+1}^k=\mathcal{W}(s_t,a_t^k,g).
\end{equation}
The resulting future state can take either a visual form (\emph{i.e.}, a GUI screenshot) or a textual form (\emph{i.e.}, renderable HTML), typically generated through diffusion-based image synthesis~\citep{vimo} or autoregressive code generation~\citep{code2world}, respectively. These forms offer complementary strengths: the former directly captures visual layout and spatial cues, whereas the latter additionally preserves structured information (\emph{e.g.}, element semantics, attributes, and coordinates).

\paragraph{Stage 2: World-Model-Guided Action Selection.}
For action selection, the simulated future states $\hat{s}_{t+1}^k$ predicted by the world model are subsequently fed to a separate scorer $\pi$ to assess the anticipated outcome of each candidate action with respect to the task goal $g$. Specifically, the scorer assigns a score $q_t^k$ to each candidate action $a_t^k$ and selects the optimal action $a_t^*$ at GUI state $s_t$:
\begin{equation}
    q_t^k = \pi(s_t, g, a_t^k, \hat{s}_{t+1}^k),
    \qquad
    a_t^* = \operatorname*{arg\,max}_{a_t^k \in \mathcal{A}_t} q_t^k.
\end{equation}
This straightforward ``Render-and-Re-Perceive'' paradigm provides the agent with explicit foresight by comparing the predicted pixel-space outcomes of candidate actions before direct execution.

\subsubsection{Model Architecture of the Proposed Method}

\paragraph{GUI-Agent Candidate Generation.}
At each interaction step, we first obtain a set of $K$ candidate actions from the frozen base GUI agent. Given the current screenshot $s_t$, task goal $g$, and interaction history $H_t$, the agent produces $\mathcal{A}_t=\{a_t^1,\ldots,a_t^K\}$, which serves as the common input to the subsequent world-model-based action evaluation. We use $K=3$ candidates proposed by the base GUI agent. This preserves the original candidate distribution. During training, the ground-truth action is not inserted into the candidate set; the target is the index of the candidate that matches the ground-truth action under the step-level matcher, and samples in which no candidate matches are retained with the agent's top-ranked candidate as the target, so that the model also learns to keep the agent's decision when no alternative is better. Duplicate candidates are not removed during training; at inference, the model is invoked only when the two top-ranked candidates differ. The same candidate-generation protocol is used at training and evaluation. For fair comparison, for a given base GUI agent and interaction step, the identical candidate-action set is provided to all compared world-model methods; no method performs method-specific candidate regeneration.

\paragraph{Shared Latent Representations.}
Our latent world-action model is instantiated with Qwen3-VL-8B-Instruct as the shared multimodal backbone. The screenshot, goal, the last six actions of the history, and all $K$ candidate actions are placed in a single prompt, and the model encodes them in one forward pass. We take the last-layer hidden state at the final prompt token as the shared representation $h_t$, yielding a representation of dimension 4096. The same representation is provided to both the action-selection and transition-prediction branches, allowing predictive transition supervision to shape the features used for candidate evaluation. The backbone is adapted with LoRA (rank 16) on all attention and MLP projections, while the original model parameters remain frozen.

\paragraph{Action Head.}
The action head maps $h_t$ to one score per candidate. We implement the head as the backbone's language-modeling output layer restricted to the $K$ index tokens (``0'', ``1'', ``2''), which amounts to a linear map with input dimension 4096 and output dimension $K$. The scores over the $K$ candidates are normalized with a softmax and supervised using the one-hot target corresponding to the annotated action; the candidate order is randomly permuted for every training example. At inference, this head provides the candidate scores used for rescoring and action selection.

\paragraph{Transition Head.}
The transition head operates on the same representation $h_t$ used by the action head and predicts the latent representation of the next state produced by the executed action, $\hat{z}_{t+1}$. We implement the transition head as a single linear layer ($4096 \rightarrow 4096$), with an output dimension matched to that of the target state representation. The transition head is used only during training to provide predictive supervision for latent dynamics and is removed at inference.

\paragraph{Target Encoder.}
To construct supervision for latent transition prediction, a target encoder $\bar{\phi}$ maps the observed GUI screenshots to latent state representations. Specifically, the next screenshot is encoded as $z_{t+1}=\bar{\phi}(s_{t+1})$, which serves as the positive target for $\hat{z}_{t+1}$, while the current screenshot is encoded as $z_t=\bar{\phi}(s_t)$ and used as an explicit negative in the contrastive objective. We use the original Qwen3-VL-8B-Instruct model without LoRA as $\bar{\phi}$, feeding the screenshot alone at up to $448\times896$ pixels, and extract its representation by mean-pooling the last-layer hidden states over non-padding tokens (4096 dimensions). The target encoder is frozen to ensure stable predictive supervision, and all target representations are precomputed once before training. Before computing the contrastive objective, all target and predicted representations are $\ell_2$-normalized.

\paragraph{Contrastive Memory Bank.}
In addition to the current-state negative $z_t$, we maintain a first-in-first-out (FIFO) memory bank $\mathcal{B}$ containing latent representations from other observed transitions. The bank provides competing future states as additional negatives, preventing the transition objective from being satisfied merely by moving the prediction away from the current state. The memory bank stores the $(z_{t+1}, z_t)$ pairs of the 192 most recent training examples (384 representations) and is updated after every training example by enqueuing that example's pair and dequeuing the oldest pair once the bank exceeds its capacity. At each optimization step, all bank entries are included as contrastive negatives. The bank is initialized empty and fills during the first 192 examples, and no deduplication is applied to near-duplicate transitions during joint training.

\subsubsection{Joint Training and Inference}

\paragraph{Training.}
We jointly optimize the action-selection and contrastive transition objectives as follows:
\begin{equation}
    \mathcal{L}
    =
    \mathcal{L}_{\mathrm{Action}}
    +
    \lambda\,\mathcal{L}_{\mathrm{Transition}},
\end{equation}
where $\lambda$ controls the contribution of the transition objective. Both objectives update the shared LoRA-adapted backbone, while each objective supervises its head. Accordingly, $\mathcal{L}_{\mathrm{Action}}$ encourages the shared representations to discriminate among candidate actions, whereas $\mathcal{L}_{\mathrm{Transition}}$ incorporates action-conditioned future information into the representation space. Their joint optimization connects predictive transition modeling with the representations used for action selection.

\paragraph{Inference.}
The transition head serves only as a training-time branch and is discarded at inference. Candidate-action scores are obtained directly from the shared backbone and action head, without generating or re-perceiving GUI screens. Thus, predictive transition supervision shapes the action-scoring representations during training without requiring transition prediction at inference.

We further adopt the confidence-based gating strategy used in prior GUI world models~\citep{code2world,vimo,xu}. When the base GUI agent assigns sufficiently high confidence to its proposed action, the latent world-action model is bypassed; otherwise, it is invoked to rescore the candidate actions. This selective invocation reduces inference overhead by applying latent world-model rescoring only to low-confidence decisions.

\subsubsection{Hyperparameters}
Table~\ref{tab:hyper} lists the hyperparameters used for training and inference in all settings.

\begin{table}[h]
\centering
\small
\begin{tabular}{ll}
\toprule
\textbf{Computing Infrastructure} & 1 NVIDIA RTX A6000 48GB GPU \\
\bottomrule
\end{tabular}

\vspace{6pt}

\begin{tabular}{cc}
\toprule
\textbf{Hyperparameter} & \textbf{Assignment} \\
\midrule
backbone & Qwen3-VL-8B-Instruct \\ \hline
precision & bfloat16 \\ \hline
LoRA rank / $\alpha$ / dropout & 16 / 32 / 0.05 \\ \hline
LoRA target modules & q, k, v, o, gate, up, down \\ \hline
trainable parameters & 43.6M (0.50\%) \\ \hline
max image pixels & 200{,}704 \\ \hline
number of candidates $K$ & 3 \\ \hline
action history length & 6 \\ \hline
transition head & linear, 4096 $\rightarrow$ 4096 \\ \hline
target encoder & frozen Qwen3-VL-8B (mean-pooled) \\ \hline
contrastive temperature $\tau$ & 0.07 \\ \hline
transition loss weight $\lambda$ & 1.0 \\ \hline
memory bank size & 192 \\ \hline
optimizer & AdamW \\ \hline
learning rate & 4e--5 \\ \hline
learning rate scheduler & constant \\ \hline
number of warmup steps & 0 \\ \hline
weight decay & 0.0 \\ \hline
effective batch size & 8 \\ \hline
number of epochs & 3 \\ \hline
training samples & 14{,}276 \\ \hline
training time & 8 h on one GPU \\ \hline
confidence threshold $c_{\mathrm{hi}}$ & 0.95 (online) / 1.0 (offline) \\ \hline
random seed & 3 \\
\bottomrule
\end{tabular}
\caption{Hyperparameters for training and inference in all settings.}
\label{tab:hyper}
\end{table}

\subsubsection{Analysis Settings}
\begin{wrapfigure}{r}{0.56\textwidth}
    \centering
    \includegraphics[width=\linewidth]{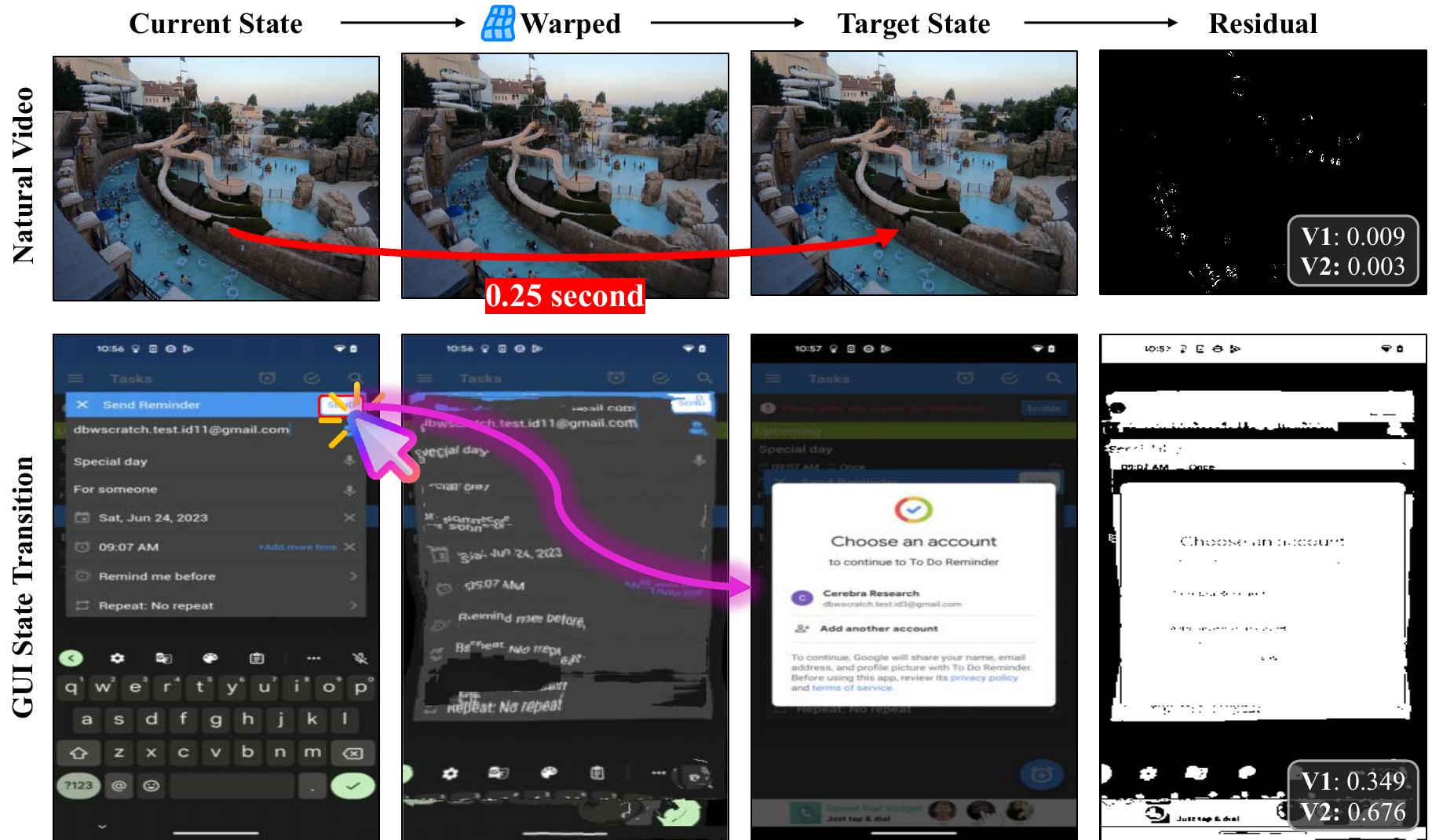}
    \caption{Natural video transitions are largely explained by pixel transport, resulting in small post-warp residuals. In contrast, GUI transitions induced by actions exhibit substantially larger residuals, revealing the inherently discontinuous nature of GUI dynamics.}
    \label{fig:sup:warp}
\end{wrapfigure}

\paragraph{GUI Transition Analysis.}
To further characterize the appearance changes observed in GUI transitions, we analyze the extent to which consecutive states can be explained by transporting existing visual content. This analysis complements Figure~\ref{fig:f2} by quantifying both the magnitude and spatial extent of pixel-level changes across GUI transitions relative to other visual domains. This perspective is particularly relevant to GUI world modeling because a single action can introduce substantial changes in screen appearance, while action selection may depend on only a small subset of the resulting information. For example, in Figure~\ref{fig:sup:warp}, selecting the ``Send'' button redirects the user to a login page; recognizing the resulting authentication requirement can be sufficient for action evaluation without reconstructing the full visual appearance of the next screen. Such transitions require explicit future-screen prediction to accommodate a broad range of appearance-level variations, motivating our focus on action-relevant future information in latent space.

We quantify these appearance changes using two complementary measures: \textbf{(V1) Post-Warp Residual Magnitude}, which measures the remaining appearance difference after transporting pixels from the current state, and \textbf{(V2) Post-Warp Residual Coverage}, which measures how spatially extensive these residual differences are. We estimate backward optical flow from the next state to the current state using RAFT-Large~\citep{raft} with its default pretrained weights and preprocessing, and warp the current RGB image into the next-state coordinate frame using bilinear interpolation. Images are resized while preserving aspect ratio such that the longer side is at most 512 pixels and are padded to dimensions divisible by eight. For each valid warped pixel, we compute the RGB L1 residual by averaging the absolute difference across the three color channels. V1 is the mean residual over valid pixels for each transition, with each video or episode summarized by the median across its transitions. V2 is the fraction of valid pixels whose residual exceeds threshold $\tau$, evaluated at 15 uniformly spaced thresholds from $0.02$ to $0.30$; transition-level coverage curves are averaged within each video or episode and summarized by their normalized area under the threshold curve.

\begin{figure}[!t]
    \centering
    \includegraphics[width=\linewidth]{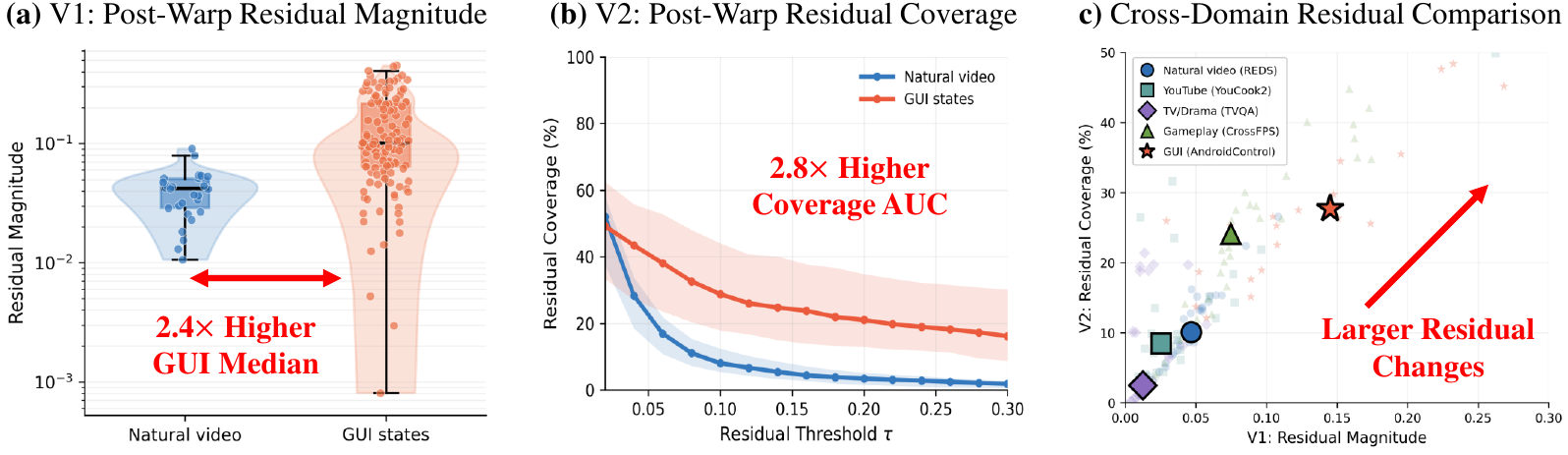}
    \caption{\textbf{(a)} GUI transitions exhibit stronger discontinuities than natural videos. \textbf{(b)} These GUI discontinuities also extend over broader spatial regions. \textbf{(c)} Across diverse visual domains, GUI dynamics show the highest degree of discontinuity, motivating latent-space GUI state forecasting.} 
    \label{fig:sup:guitransition}
\end{figure}
For the comparison with natural video, we use the complete AndroidControl validation split containing 137 episodes and 690 action-conditioned transitions, together with all 30 REDS validation videos. REDS frame pairs are sampled at 0.25-second intervals, whereas GUI pairs correspond directly to observed current- and next-state transitions. As shown in Figure~\ref{fig:sup:guitransition}(a) and (b), GUI transitions exhibit $2.4\times$ larger residual magnitude and $2.8\times$ higher residual-coverage AUC than natural videos~\citep{reds}. For the cross-domain comparison in Figure~\ref{fig:sup:guitransition}(c), we evaluate REDS, YouCook2~\citep{zhou}, TVQA~\citep{tvqa}, CrossFPS~\citep{scope}, and AndroidControl using a common protocol of 30 units per domain and three transitions per unit. Video pairs are separated by one-third of a second and sampled at three positions distributed across each clip, while three GUI transitions are selected evenly within each episode. Dataset selection is deterministic with seed 42. Overall, GUI transitions exhibit larger and more spatially extensive appearance changes than the other evaluated domains. These observations support the main-paper motivation that pixel-space GUI world models must accommodate a broad range of future appearances, even though action selection may require only the action-relevant consequence of those transitions.

\paragraph{Identity-Shortcut Analysis.}
For the analysis in Figure~\ref{fig:f4}, we randomly sample 2.5K transitions from AndroidControl and evaluate the pixel-space world models according to the magnitude of the ground-truth GUI transition. We first convert the current and next screenshots to grayscale and resize them to $64\times64$. For a transition $(s_t,s_{t+1})$, we define its transition magnitude as the mean absolute pixel difference between the two processed screenshots. We divide the samples into weak, medium, and strong transitions using thresholds of $0.08$ and $0.24$: weak transitions have magnitude below $0.08$, medium transitions fall between $0.08$ and $0.24$, and strong transitions have magnitude at least $0.24$. To identify an identity shortcut, we additionally compare the predicted future screen $\hat{s}_{t+1}$ with the current screen $s_t$ using the same pixel-change measure. A prediction is classified as an identity shortcut when the ground-truth transition is larger than the predicted transition, $d(s_t, s_{t+1}) > d(s_t, \hat{s}_{t+1})$, and the prediction remains close to the current screen, $d(s_t, \hat{s}_{t+1}) < 0.08$, where $d(\cdot,\cdot)$ denotes the same mean absolute pixel difference. Shortcut severity is computed as the fraction of predictions satisfying this criterion within each transition-magnitude group. We report the downstream step-level success rate on the same samples across methods to jointly examine shortcut behavior and action-selection performance across transition magnitudes.

\paragraph{Action-Specific Representation Analysis.}
For the representation analysis in Figure~\ref{fig:f5}, we sample 900 transitions from the AndroidControl test set, stratified over action types (300 each for click, scroll, and text input). Because current GUI states can exhibit correlated action distributions, applying dimensionality reduction to the predicted representations may confound current-state similarity with action-induced structure. We therefore cluster the current-state representations $z_t$ using $k$-means with 100 clusters, and keep the 52 clusters that contain at least six samples and at least two action types. For each sample, we compute a residual representation by subtracting the mean predicted representation of samples belonging to the same current-state cluster from its predicted future representation. We then apply t-SNE to these residual representations and visualize the embeddings by action type. We use the frozen Qwen3-VL-8B target encoder (last-layer states mean-pooled over non-padding tokens) to embed the current screen for the base agent and the rendered next screen for the pixel-space world models, and the transition-head output at the last token of the ground-truth candidate in a three-candidate prompt for our model, with t-SNE perplexity 30, learning rate \texttt{auto}, 1{,}000 iterations, and random seed 0. Residual representations are $\ell_2$-normalized before t-SNE. This probe reads the representation at the ground-truth candidate's tokens rather than at the final prompt token used for training and inference; the final-token representation, which predicts the next state of the annotated action, does not separate action types under the same residual analysis.

\subsection{Dataset Details}

\paragraph{AndroidControl.}
AndroidControl~\citep{androidcontrol} is a large-scale mobile GUI interaction dataset containing 15,283 human demonstrations over 14,548 unique tasks and 833 Android applications. Each trajectory provides screenshots, accessibility-tree observations, natural-language instructions, and corresponding interaction actions. A distinctive feature of AndroidControl is that each task includes both high-level goal instructions and low-level step-wise instructions, supporting evaluation of GUI agents under different levels of task abstraction. In our experiments, we use AndroidControl for offline in-distribution evaluation of step-level action selection.

\paragraph{AITZ.}
Android-In-The-Zoo (AITZ)~\citep{aitz} is a mobile GUI interaction dataset constructed from Android trajectories with detailed action-level annotations. The dataset contains 18,643 screen--action pairs and augments interaction steps with chain-of-action-thought annotations describing the reasoning associated with GUI actions. It spans diverse mobile interaction scenarios, including general application use, Google applications, installation tasks, single-app interactions, and web shopping. We use AITZ as an additional offline in-distribution benchmark to evaluate whether the proposed action-selection framework generalizes across mobile interaction datasets.

\paragraph{AndroidWorld.}
AndroidWorld~\citep{androidworld} is an interactive Android environment designed for evaluating autonomous GUI agents through online execution. It contains 116 parameterized tasks across 20 real-world Android applications and instantiates task configurations rather than relying solely on a fixed set of recorded trajectories. Unlike the offline benchmarks, the agent's predicted action is executed in the environment and determines subsequent GUI states, allowing errors and improvements in action selection to propagate throughout the trajectory. We therefore use AndroidWorld for online evaluation and measure performance based on end-to-end task completion.

\paragraph{GUI-Odyssey.}
GUI-Odyssey~\citep{lu2025guiodyssey} is a mobile GUI navigation dataset designed to evaluate long-horizon cross-application interaction. Its trajectories require agents to execute sequences of actions that may span multiple applications, providing a substantially different interaction distribution from the predominantly single-app settings used in our in-distribution evaluation. We use GUI-Odyssey exclusively for out-of-distribution evaluation, without adapting the proposed model to its evaluation trajectories, to assess whether the learned action-selection representations transfer to unseen cross-app navigation settings under a pronounced distribution shift.

\subsection{Evaluation Details}
This section provides additional details on the evaluation protocols and metrics used for offline, online, out-of-distribution, and efficiency evaluation.

\subsubsection{Offline Evaluation}

\paragraph{Step-Level Evaluation Metrics.}
For offline evaluation on AndroidControl and AITZ, we evaluate each interaction step independently using three metrics: action type accuracy (Type), grounding accuracy (Grounding), and step-level success rate (SR). Type measures whether the predicted action type matches the annotated action type. Grounding is evaluated only for click actions and is considered correct when the predicted coordinate falls within a radius corresponding to 14\% of the screen width from the annotated target point. SR requires both the action type and its associated argument to be correct. Specifically, a click action is successful when the predicted coordinate lies within the target element's bounding box, a text-input action when the predicted text matches the annotated content, and a scroll action when the predicted direction matches the annotation. All methods are evaluated on the same examples and candidate-action sets for each base GUI agent.

\subsubsection{Online Evaluation}

\paragraph{Task-Level Evaluation Protocol.}
For online evaluation, we use AndroidWorld, where the agent's selected actions are executed directly in the interactive environment and therefore determine subsequent GUI states throughout the trajectory. We report task success rate (\textbf{SR}) according to the benchmark's task-completion criterion, separately for Easy, Medium, and Hard tasks as well as overall. To characterize interaction efficiency, we additionally report the average number of interaction steps (Mean Steps), total token usage per task (Tokens/Task), and execution time (Run Time). Each task is evaluated using a single task instance per task template (116 templates), with task parameters sampled from a fixed seed (30) so that every compared method receives identical goals, and with a maximum interaction budget of $10\times$ the task's complexity score as defined by AndroidWorld. Before each rollout, the environment is reset by the task's own initialization routine, which restores the required app states and returns the device to the home screen, and each rollout is followed by the corresponding tear-down routine, ensuring consistent initial conditions across compared methods.

\subsubsection{Out-of-Distribution Evaluation}

\paragraph{Evaluation Protocol.}
We evaluate out-of-distribution generalization on GUI-Odyssey using the same step-level Type, Grounding, and SR metrics defined for the offline evaluation. GUI-Odyssey contains cross-application and longer-horizon interaction trajectories that differ from the in-distribution AndroidControl and AITZ settings, allowing us to examine whether the learned action-selection representations transfer beyond the training distribution under long-horizon cross-app navigation conditions. We evaluate GUI-Odyssey using 160 episodes (1{,}598 steps) drawn from the test set of its official random split: the test episodes are shuffled with a fixed seed, episodes with 4 to 12 steps whose actions map to our action space are kept, and the first 160 are used. We apply the same candidate-generation and rescoring protocol across all compared methods. The model is evaluated without any GUI-Odyssey fine-tuning or adaptation; it is trained only on AndroidControl and AITZ, and no GUI-Odyssey episode is used for training, validation, or hyperparameter selection.

\subsubsection{Efficiency Evaluation}
We evaluate computational efficiency using Tokens/Task, Mean Steps, and Run Time, depending on the evaluation setting. Tokens/Task measures the total number of input and output tokens, including image tokens, consumed by all model calls made for a task: every action-selection and summary call of the base GUI agent, the generation calls of language-model-based world models (the Code2World prompt and generated HTML), and the prompt of the scoring model (our judge or the world-model selector) on the steps where it is invoked; for API backbones, the token counts reported by the provider are used, and in the offline setting the same quantity is reported per step. Mean Steps, reported for online evaluation, denotes the average number of environment interactions taken per task. Run Time measures the end-to-end wall-clock time in seconds per task, measured from task initialization to the end of the episode and including agent inference, world-model generation or rendering, scoring, and environment execution; in the offline setting, it is the per-step sum of the proposer, world-model, and scorer latencies. For diffusion-based world models, token usage associated with image generation is not applicable and is therefore omitted. For methods using confidence-based gating, the reported efficiency reflects the same selective world-model invocation policy used for the corresponding performance evaluation under deployment conditions.

\subsection{Pseudocode}
\newcounter{algline}
\newcommand{\algreset}{\setcounter{algline}{0}}
\newcommand{\algl}[2][0]{\stepcounter{algline}%
  \makebox[1.5em][r]{\footnotesize\arabic{algline}:}\hspace{0.7em}\hspace*{#1em}#2\par}
\newcommand{\algtitle}[2]{\hrule\vspace{2pt}{\normalsize\textbf{Algorithm #1} #2}\vspace{2pt}\hrule\vspace{3pt}}

\begin{table}[h]
\small\setlength{\parindent}{0pt}
\algreset
\algtitle{1}{Training the latent transition judge}
\algl{\textbf{Require:} frozen VLM $f_\theta$ with LoRA $\phi$, transition head $A$, frozen target encoder $E$, dataset $\mathcal{D}$, temperature $\tau$, weight $\lambda$, bank size $M$}
\algl{Initialise FIFO bank $\mathcal{B} \gets \emptyset$}
\algl{\textbf{while} not converge \textbf{do}}
\algl[1.5]{Choose $(s_t, g, H_t, \{a_k\}_{k=1}^{K}, y, s_{t+1})$ from $\mathcal{D}$}
\algl[1.5]{Shuffle candidate order; relabel $y$ accordingly}
\algl[1.5]{\# Action selection}
\algl[1.5]{$x \gets \mathrm{Prompt}(s_t, g, H_t, \{a_k\}_{k=1}^{K})$}
\algl[1.5]{$\ell_{0:K-1},\, h \gets f_{\theta,\phi}(x)$ \hfill (index-token logits, last-token state)}
\algl[1.5]{$\mathcal{L}_{\mathrm{pick}} \gets -\log \mathrm{softmax}(\ell)_{y}$}
\algl[1.5]{\# Transition contrast}
\algl[1.5]{$e \gets \mathrm{normalise}(A(h))$}
\algl[1.5]{$z_{t+1} \gets E(s_{t+1})$, $\; z_t \gets E(s_t)$}
\algl[1.5]{$\mathcal{N} \gets \{z_t\} \cup \mathcal{B}$ \hfill (current screen and bank as negatives)}
\algl[1.5]{$\mathcal{L}_{\mathrm{trans}} \gets -\log \dfrac{\exp(\mathrm{sim}(e, z_{t+1})/\tau)}{\exp(\mathrm{sim}(e, z_{t+1})/\tau) + \sum_{z' \in \mathcal{N}} \exp(\mathrm{sim}(e, z')/\tau)}$}
\algl[1.5]{Take gradient step on $\nabla_{\phi, A}\,(\mathcal{L}_{\mathrm{pick}} + \lambda \mathcal{L}_{\mathrm{trans}})$}
\algl[1.5]{Push $(z_{t+1}, z_t)$ into $\mathcal{B}$; drop the oldest pair if $|\mathcal{B}| > M$}
\algl{\textbf{end while}}
\vspace{2pt}\hrule
\end{table}

\begin{table}[h]
\small\setlength{\parindent}{0pt}
\algreset
\algtitle{2}{Gated action selection at inference}
\algl{\textbf{Require:} frozen agent $\pi$, trained judge $f_{\theta,\phi}$, confidence threshold $c_{\mathrm{hi}}$}
\algl{\textbf{while} task not finished \textbf{do}}
\algl[1.5]{$\{(a_k, c_k)\}_{k=1}^{K} \gets \pi(s_t, g, H_t)$ \hfill (candidates with confidences, most likely first)}
\algl[1.5]{\# Gate}
\algl[1.5]{\textbf{if} fewer than $K$ candidates \textbf{or} $a_1$ is terminal \textbf{or} $a_1 = a_2$ \textbf{or} $|\{k : c_k \ge c_{\mathrm{hi}}\}| = 1$ \textbf{then}}
\algl[3]{$\hat a \gets a_1$}
\algl[1.5]{\textbf{else}}
\algl[3]{\# Judge (one forward pass, transition head unused)}
\algl[3]{$x \gets \mathrm{Prompt}(s_t, g, H_t, \{a_k\}_{k=1}^{K})$}
\algl[3]{$\ell_{0:K-1} \gets f_{\theta,\phi}(x)$}
\algl[3]{$\hat a \gets a_{\arg\max_k \ell_k + 1}$}
\algl[1.5]{\textbf{end if}}
\algl[1.5]{Execute $\hat a$; $H_{t+1} \gets H_t \cup \{\hat a\}$; observe $s_{t+1}$}
\algl{\textbf{end while}}
\vspace{2pt}\hrule
\end{table}









\clearpage
\section{Additional Qualitative Results}

\subsection{GUI Navigation Trajectories}
\begin{figure*}[h]\centering
\includegraphics[width=\linewidth]{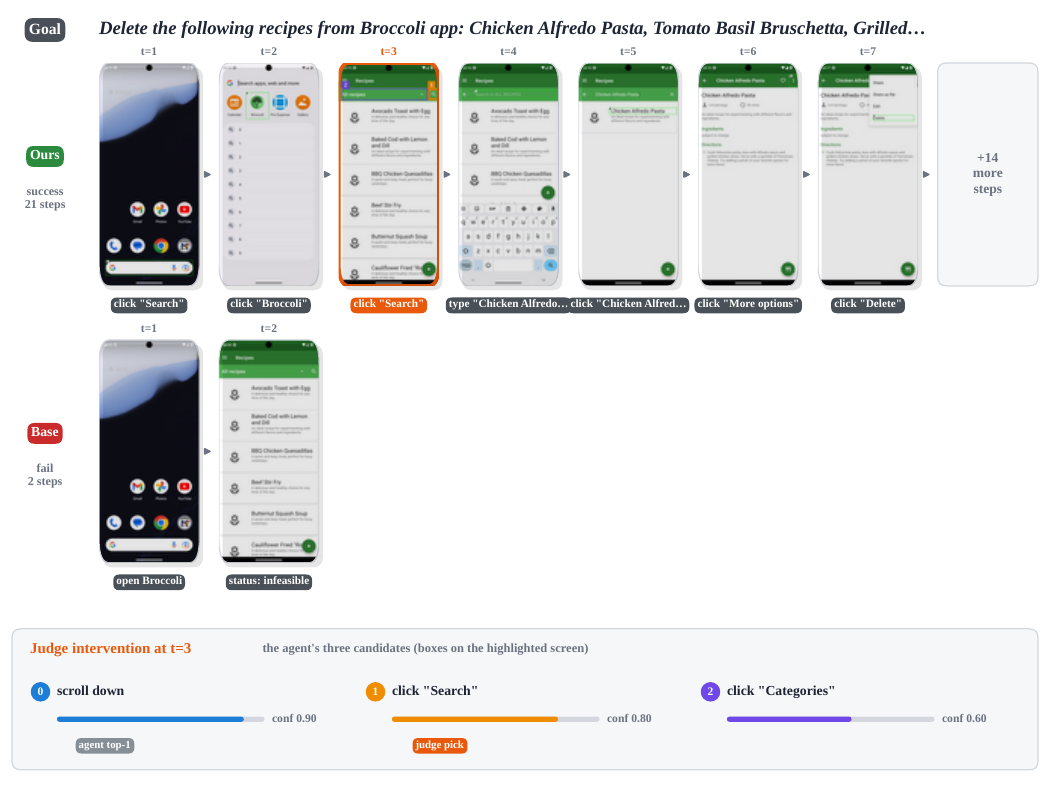}
\caption{\textbf{Long-horizon recovery.} At step 3 the agent prefers to scroll the full recipe list; our model selects the search button instead. The agent then finds and deletes each recipe and completes the 21-step task, whereas Base declares the task infeasible at step 2.}
\label{fig:traj_recipe}
\end{figure*}

\begin{figure*}[h]\centering
\includegraphics[width=\linewidth]{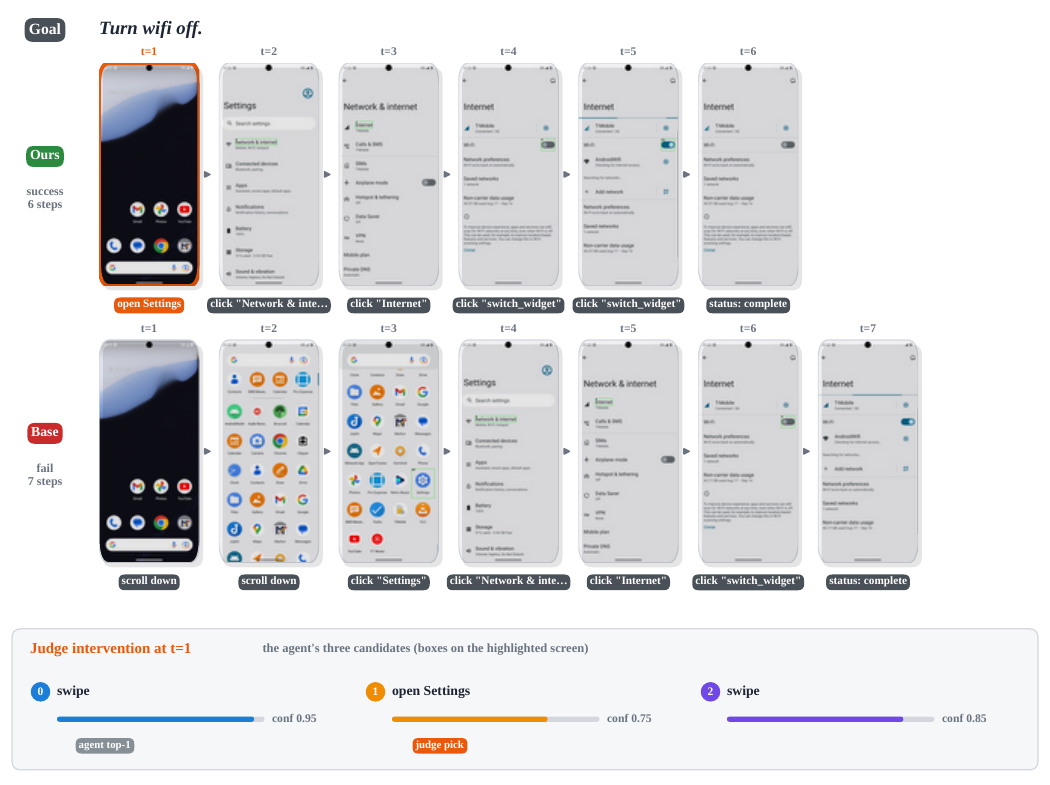}
\caption{\textbf{Same intervention, different task.} After the same first-step correction, Ours turns Wi-Fi off and verifies the toggle state. Base reaches the same toggle but leaves Wi-Fi on before declaring completion.}
\label{fig:traj_wifi}
\end{figure*}

Figures~\ref{fig:traj_recipe}--\ref{fig:traj_wifi} show complete AndroidWorld episodes of the frozen Qwen3-VL-8B agent with our latent world-action model (Ours, top row) and without it (Base, bottom row), on the same task instance and seed. Every screen is the state actually reached in the emulator. The orange frame marks the step at which the gate opened and our model replaced the agent's top-ranked candidate; the numbered boxes on that screen are the agent's three candidates, and the bottom panel lists them with the agent's own confidences to facilitate direct step-by-step qualitative comparison.

\begin{figure*}[h]\centering
\includegraphics[width=\linewidth]{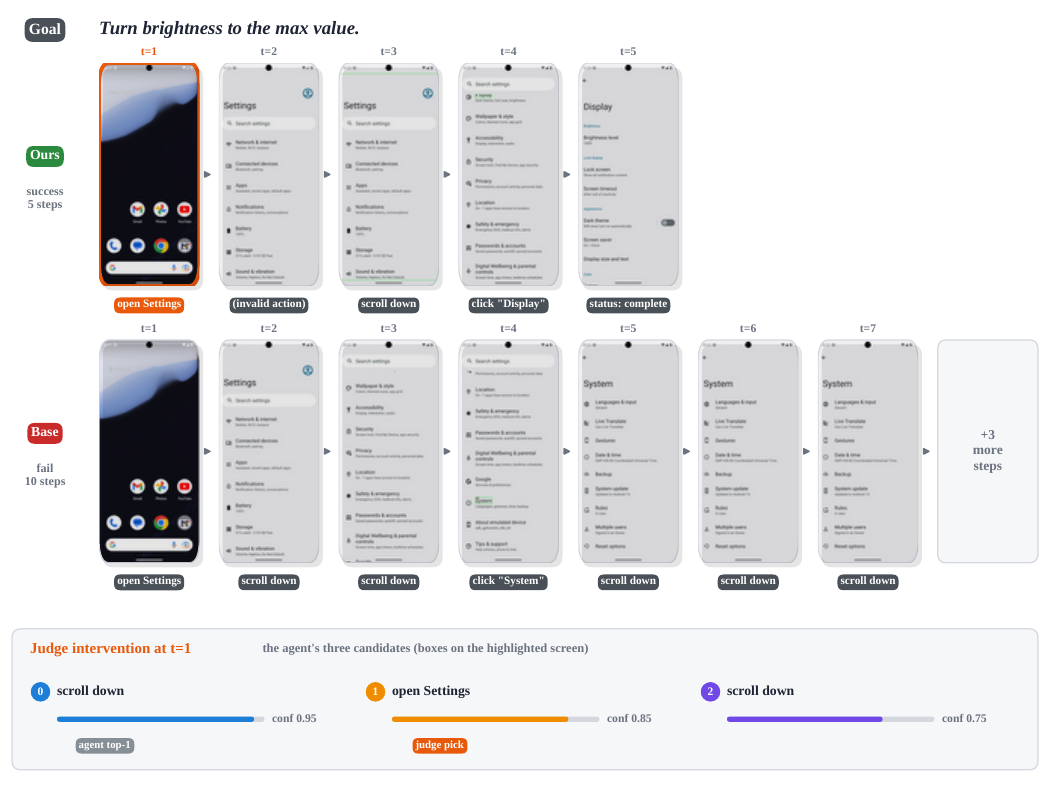}
\caption{\textbf{Choosing a reliable route.} The agent's top candidate is a swipe that opens quick settings; our model selects \emph{open Settings}, and the agent reaches the brightness control in five steps. Base keeps scrolling within the System page until its step budget runs out.}
\end{figure*}

\clearpage
\subsection{Additional Identity-Shortcut Examples}

\begin{figure*}[h]\centering
\includegraphics[width=\linewidth]{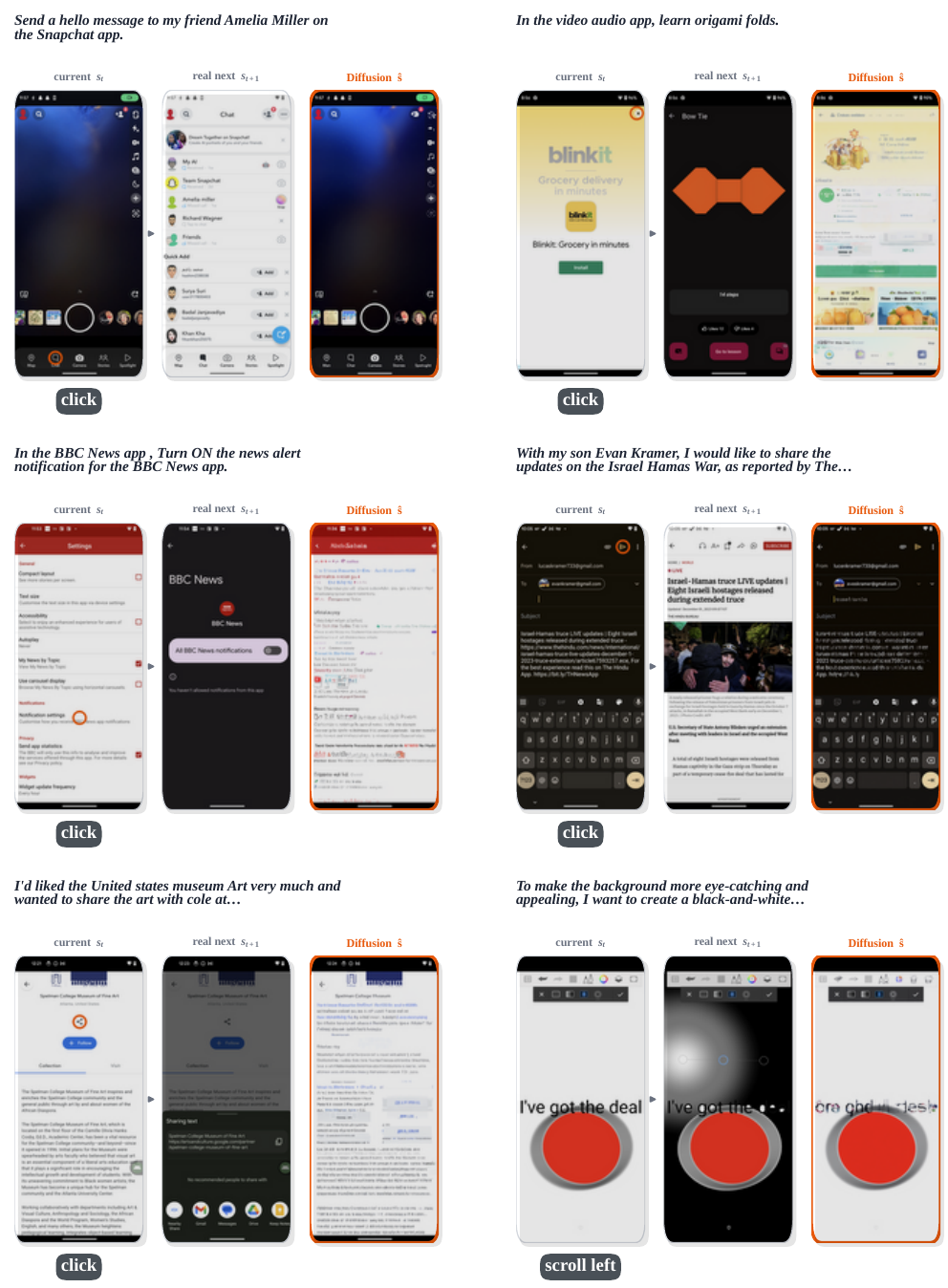}
\caption{\textbf{Identity shortcut of the diffusion-based world model (1/2).}}
\label{fig:sc_diff1}
\end{figure*}

\begin{figure*}[h]\centering
\includegraphics[width=\linewidth]{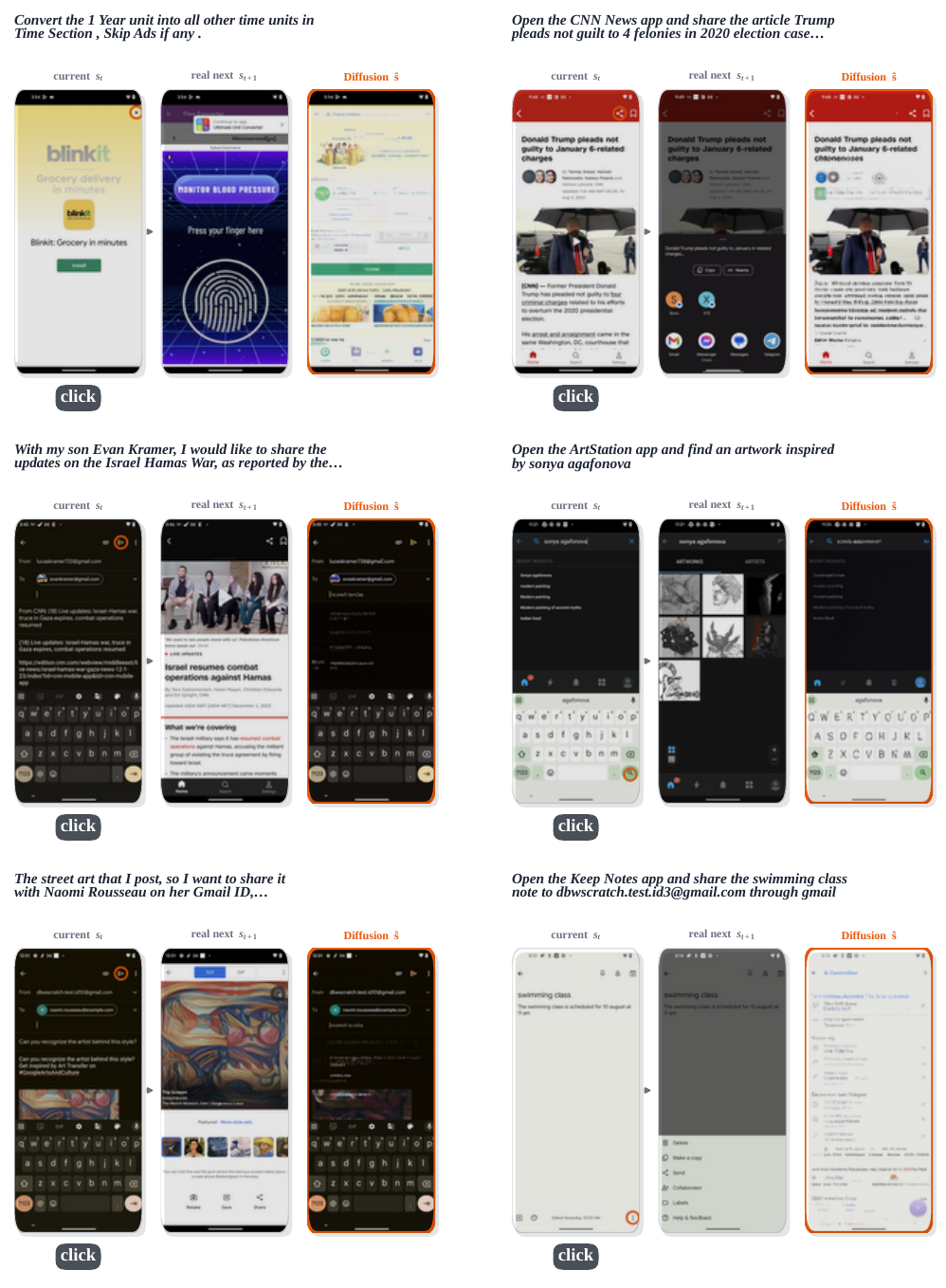}
\caption{\textbf{Identity shortcut of the diffusion-based world model (2/2).}}
\label{fig:sc_diff2}
\end{figure*}

\paragraph{Diffusion-Based World Model.}
Across all transition magnitudes, the diffusion model reproduces the current screen with only local texture changes, even when the real next screen belongs to a different page or application (Figures~\ref{fig:sc_diff1}--\ref{fig:sc_diff2}).

\begin{figure*}[h]\centering
\includegraphics[width=\linewidth]{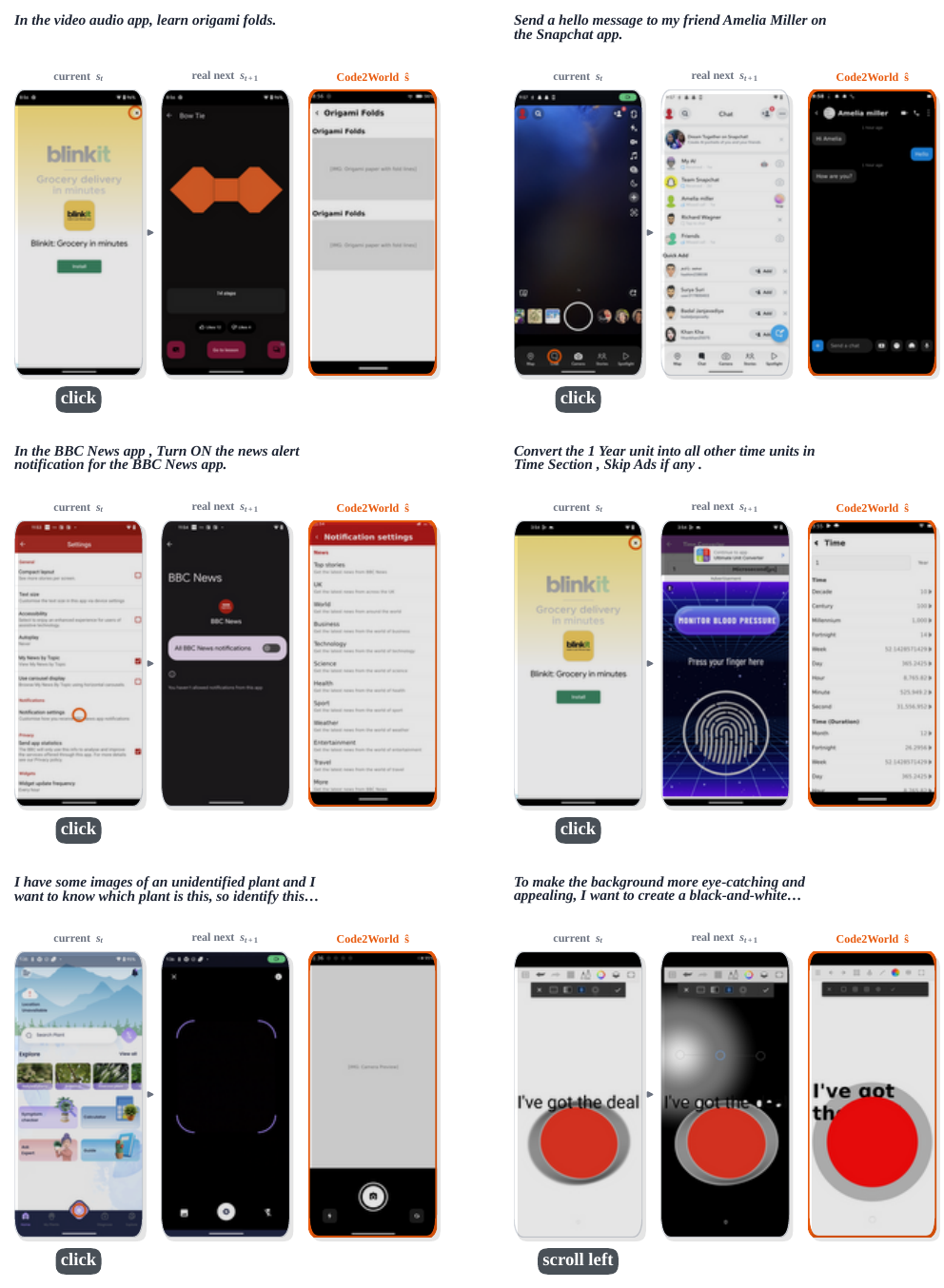}
\caption{\textbf{Code2World predictions on transitions with large real screen changes (1/2).}}
\label{fig:sc_c2w1}
\end{figure*}

\begin{figure*}[h]\centering
\includegraphics[width=\linewidth]{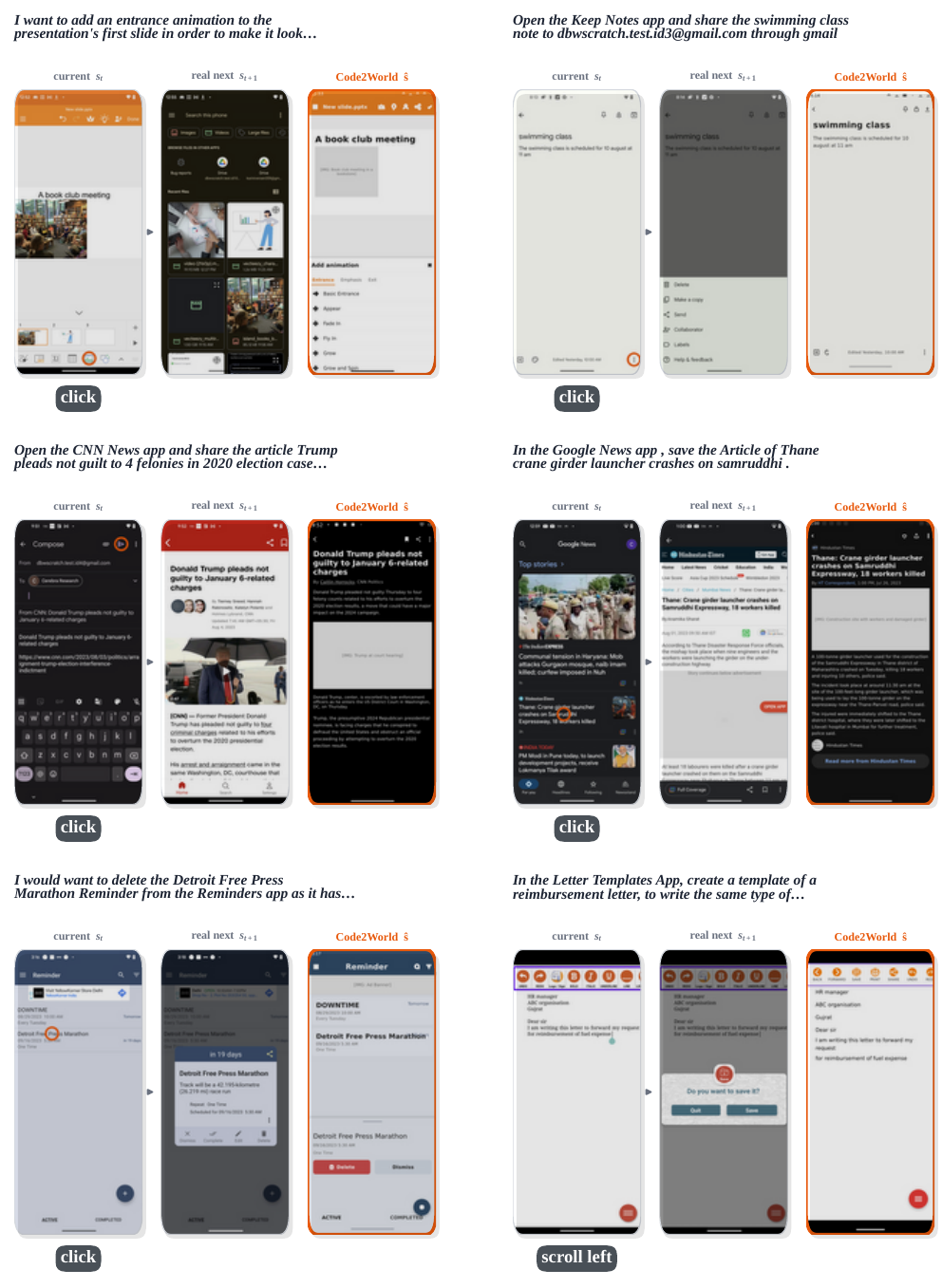}
\caption{\textbf{Code2World predictions on transitions with large real screen changes (2/2).}}
\label{fig:sc_c2w2}
\end{figure*}

\clearpage
\paragraph{Code2World.}
Because Code2World regenerates the page as HTML, its predictions rarely copy the current screen pixel for pixel. Its failures are of a different kind: the rendered page is well formed and plausible for the application, but it does not correspond to the real outcome of the action (Figures~\ref{fig:sc_c2w1}--\ref{fig:sc_c2w2}).

\clearpage
\section{Prompts}
\label{app:prompts}
All prompts are reproduced from the released code; curly braces mark fields filled at run time.
The agent prompt is the unmodified M3A prompt of AndroidWorld. Our method only appends the candidate-generation suffix (Prompt~\ref{prompt:cand_online}) and consults the judge with the scoring prompt (Prompt~\ref{prompt:score}).

\subsection{GUI-Agent Prompt}

\begin{figure*}[ht!]
\begin{tcolorbox}[promptstyle, title=M3A Action-Selection Prompt]
\scriptsize
You are an agent who can operate an Android phone on behalf of a user. Based on user's goal/request, you may\\
- Answer back if the request/goal is a question (or a chat message), like user asks ``What is my schedule for today?''.\\
- Complete some tasks described in the requests/goals by performing actions (step by step) on the phone.\par\medskip

When given a user request, you will try to complete it step by step. At each step, you will be given the current screenshot (including the original screenshot and the same screenshot with bounding boxes and numeric indexes added to some UI elements) and a history of what you have done (in text). Based on these pieces of information and the goal, you must choose to perform one of the action in the following list (action description followed by the JSON format) by outputing the action in the correct JSON format.\par\medskip

\textbf{ACTIONS:}\\
- Task completed: \{``action\_type'': ``status'', ``goal\_status'': ``complete''\}\\
- Task infeasible: \{``action\_type'': ``status'', ``goal\_status'': ``infeasible''\}\\
- Answer user's question: \{``action\_type'': ``answer'', ``text'': $<$answer\_text$>$\}\\
- Click/tap on an element (use the numeric index on the bounding box): \{``action\_type'': ``click'', ``index'': $<$target\_index$>$\}\\
- Long press on an element: \{``action\_type'': ``long\_press'', ``index'': $<$target\_index$>$\}\\
- Type text into a text field (includes clicking the field and pressing enter): \{``action\_type'': ``input\_text'', ``text'': $<$text\_input$>$, ``index'': $<$target\_index$>$\}\\
- Press the Enter key: \{``action\_type'': ``keyboard\_enter''\}\\
- Navigate to the home screen: \{``action\_type'': ``navigate\_home''\}\\
- Navigate back: \{``action\_type'': ``navigate\_back''\}\\
- Scroll the screen or a scrollable UI element: \{``action\_type'': ``scroll'', ``direction'': $<$up, down, left, right$>$, ``index'': $<$optional\_target\_index$>$\}\\
- Open an app: \{``action\_type'': ``open\_app'', ``app\_name'': $<$name$>$\}\\
- Wait for the screen to update: \{``action\_type'': ``wait''\}\par\medskip

\textbf{INPUT:}\\
The current user goal/request is: \{goal\}\\
Here is a history of what you have done so far: \{history\}\\
The current screenshot and the same screenshot with bounding boxes and labels added are also given to you.\\
Here is a list of detailed information for some of the UI elements (notice that some elements in this list may not be visible in the current screen and so you can not interact with it, can try to scroll the screen to reveal it first), the numeric indexes are consistent with the ones in the labeled screenshot: \{ui\_elements\}\par\medskip

\textbf{GUIDELINES:}\\
General / Action Related / Text Related Operations (the unmodified M3A guidelines; omitted for space).\par\medskip

\textbf{OUTPUT:}\\
Now output an action from the above list in the correct JSON format, following the reason why you do that. Your answer should look like:\\
Reason: ...\\
Action: \{``action\_type'':...\}
\end{tcolorbox}
\label{prompt:agent}
\end{figure*}

\clearpage
\subsection{Candidate-Generation Prompt}

\begin{figure*}[ht!]
\begin{tcolorbox}[promptstyle, title=Online Candidate Suffix (appended to the M3A prompt)]
\scriptsize
\textbf{IMPORTANT --- output format for THIS step:}\\
Propose 3 DISTINCT candidate actions to try next (each a possible single next action for the CURRENT step, not a sequence), each exactly as:\\
CANDIDATE 0\\
Reason: $<$one line$>$\\
Action: \{json\}\\
Confidence: $<$0.1--1.0$>$\\
CANDIDATE 1\\
Reason: $<$one line$>$\\
Action: \{json\}\\
Confidence: $<$0.1--1.0$>$\\
CANDIDATE 2\\
Reason: $<$one line$>$\\
Action: \{json\}\\
Confidence: $<$0.1--1.0$>$\par\medskip

Order them best-first. Confidence = how likely that action is the correct next one (0.8--0.9 strong evidence, 0.6--0.7 some ambiguity, 0.1--0.3 unlikely). Use the same JSON action schema as instructed above.
\end{tcolorbox}
\label{prompt:cand_online}
\end{figure*}

\begin{figure*}[ht!]
\begin{tcolorbox}[promptstyle, title=Offline Candidate Prompt (AndroidControl / AITZ / GUI-Odyssey)]
\scriptsize
Now propose exactly 3 DISTINCT candidate actions for the NEXT step --- three alternatives for this single step, NOT a sequence --- ordered most-likely first. Do NOT output a $<$tool\_call$>$. Instead output ONLY one JSON object of exactly this form (all three keys required):\\
\{``1'': \{``Confidence'': $<$number in [0.1, 1.0]$>$, ``Action'': \{``action'': ``...'', ...\}\}, ``2'': \{``Confidence'': ..., ``Action'': \{...\}\}, ``3'': \{``Confidence'': ..., ``Action'': \{...\}\}\}\\
Each ``Action'' object uses the mobile\_use argument schema above (action, coordinate, coordinate2, text, button, status, time).\par\medskip

\textbf{CONFIDENCE SEMANTICS:}\\
1.0 = certain; 0.8--0.9 high confidence with strong evidence; 0.6--0.7 moderate with some ambiguity; 0.4--0.5 low due to significant uncertainty; 0.1--0.3 very low.\\
If you feel there are no other reasonable candidates for this step, set the confidence of the remaining candidates to 0.1.\\
Begin your reply with ``\{''.
\end{tcolorbox}
\label{prompt:cand_offline}
\end{figure*}

\clearpage
\subsection{World-Model Baseline Prompts}

\begin{figure*}[ht!]
\begin{tcolorbox}[promptstyle, title=Code2World System Prompt]
\scriptsize
You are an expert UI State Transition Simulator and Frontend Developer.\\
Your task is to predict the NEXT UI STATE based on a screenshot of the current state and a user interaction.\par\medskip

\textbf{IMAGE INTERPRETATION RULES:}\\
- Red Circle: Indicates a Click or Long Press target at that location.\\
- Red Arrow: Indicates a Scroll or Swipe. The arrow points in the direction of finger movement.\\
- These cues exist ONLY to show the action. DO NOT render these red circles or arrows in your output HTML.\par\medskip

\textbf{CRITICAL STRUCTURAL RULES:}\\
- Output ONLY raw HTML. Start with $<$!DOCTYPE html$>$ and end with $<$/html$>$.\\
- All visible content MUST be wrapped in $<$div id=``render-target''$>$ with width 1080px, height 2400px, position relative, overflow hidden. [...]\par\medskip

\textbf{CONTENT GENERATION LOGIC:}\\
- Transition: Analyze the action. If the user clicks a button, show the result (e.g., a menu opens, a checkbox checks, page navigates).\\
- Images: Use semantic text placeholders. DO NOT use real URLs.\\
- Icons: Use simple inline SVG paths or Unicode.\par\medskip

\textbf{OUTPUT REQUIREMENT:}\\
- Do NOT generate Markdown blocks. Do NOT provide explanations or conversational text. Output the code directly.
\end{tcolorbox}
\label{prompt:c2w_system}
\end{figure*}

\begin{figure*}[ht!]
\begin{tcolorbox}[promptstyle, title=Code2World User Prompt and Diffusion Conditioning Text]
\scriptsize
\textbf{CODE2WORLD USER PROMPT:}\\
$<$image$>$ (current screenshot with the action drawn as a red circle or arrow)\\
INPUT CONTEXT\\
1. User Intent: ``\{goal\}''\\
2. Interaction Details: Description: \{semantic\_desc\}; Action Data: \{action\_json\}\\
COMMAND\\
Based on the visual cues in the image and the interaction data above, generate the HTML for the RESULTING UI STATE (what the screen looks like after this action).\par\medskip

\textbf{\{semantic\_desc\} BY ACTION TYPE:}\\
- click: User performed a CLICK at coordinates (x, y). Expect the button/element at this location to trigger.\\
- long\_press: User performed a LONG PRESS at coordinates (x, y). Expect a context menu or selection state.\\
- scroll: User SCROLLED $<$DIRECTION$>$. The content should move, revealing new items from the $<$direction$>$ direction.\\
- input\_text: User is TYPING the text: `$<$text$>$'. The focused input field MUST now contain this text.\\
- open\_app: System Context Switch: The user opened the app `$<$app$>$'. Show the home screen of this app.\\
- navigate\_back: System Navigation: The user pressed BACK. Return to the previous screen.\\
- navigate\_home: System Navigation: The user pressed HOME. Show the Desktop.\\
- wait: Action: WAIT. Keep the UI mostly unchanged unless loading completes.\par\medskip

\textbf{DIFFUSION CONDITIONING TEXT:}\\
Predict the next page state via image from this current screenshot using action description ``\{desc\}'' and action target ``\{tgt\}'' and relative coordinates ``[\{x\}, \{y\}]''.\\
- click / long\_press: (``click'' / ``long press'', ``point(x, y)'')\\
- scroll: (``scroll $<$direction$>$'', ``screen'')\\
- input\_text: (``input text: $<$text$>$'', ``text field'')\\
- open\_app: (``open app: $<$app$>$'', ``app icon'')\\
- other: ($<$action\_type$>$, ``screen'')
\end{tcolorbox}
\label{prompt:wm}
\end{figure*}

\clearpage
\subsection{Evaluation and Scoring Prompts}
The same prompt text is shared by all selectors so that they differ only in the images that follow each candidate. For the Code2World and diffusion selectors, these are the rendered next screens. For our model, they are fixed blank placeholders: no world model is run and no future screen is generated or perceived at inference, and the candidate scores are read directly from the index-token logits. The phrase ``each followed by a world model's prediction of the resulting screen'' is therefore kept only to keep the prompt identical across selectors.

\begin{figure*}[ht!]
\begin{tcolorbox}[promptstyle, title=Scoring Prompt]
\scriptsize
\textbf{IMAGES:}\\
$[$image: current screenshot$]$\\
$[$image: prediction for candidate 0$]$ $[$image: prediction for candidate 1$]$ $[$image: prediction for candidate 2$]$\\
(blank placeholder for our judge; rendered next screen for the Code2World and diffusion selectors)\par\medskip

\textbf{TEXT:}\\
This is the current mobile-GUI screen (first image). Goal: ``\{goal\}''.\\
Steps already completed:\\
\{hist\}\\
Candidate next actions, each followed by a world model's prediction of the resulting screen:\\
0. \{candidate 0\}\\
1. \{candidate 1\}\\
2. \{candidate 2\}\\
Which candidate best advances the goal? Answer with ONLY the number (0--2).\par\medskip

\textbf{HISTORY FORMAT:}\\
- \{hist\} lists the last six own actions as ``Step $j$: $<$action text$>$'', or ``(none - this is the first step)''.\\
- The logits of the index tokens are read at the last position; no text is generated.
\end{tcolorbox}
\label{prompt:score}
\end{figure*}

\end{document}